# Soft Constraints of Difference and Equality

**Emmanuel Hebrard**                                    HEBRARD@LAAS.FR
*CNRS; LAAS*
*Université de Toulouse*
*Toulouse, France*

**Dániel Marx**                                         DMARX@CS.BME.HU
*Humboldt-Universität zu Berlin*
*Berlin, Germany*

**Barry O'Sullivan**                                    B.OSULLIVAN@CS.UCC.IE
*Cork Constraint Computation Centre*
*Department of Computer Science, University College Cork*
*Cork, Ireland*

**Igor Razgon**                                         IR45@MCS.LE.AC.UK
*Department of Computer Science, University of Leicester*
*Leicester, United Kingdom*

## Abstract

In many combinatorial problems one may need to model the diversity or similarity of sets of assignments. For example, one may wish to maximise or minimise the number of distinct values in a solution. To formulate problems of this type we can use soft variants of the well known ALLDIFFERENT and ALLEQUAL constraints. We present a taxonomy of six soft global constraints, generated by combining the two latter ones and the two standard cost functions, which are either maximised or minimised. We characterise the complexity of achieving arc and bounds consistency on these constraints, resolving those cases for which NP-hardness was neither proven nor disproven. In particular, we explore in depth the constraint ensuring that at least $k$ pairs of variables have a common value. We show that achieving arc consistency is NP-hard, however bounds consistency can be achieved in polynomial time through dynamic programming. Moreover, we show that the maximum number of pairs of equal variables can be approximated by a factor of $\frac{1}{2}$ with a linear time greedy algorithm. Finally, we provide a fixed parameter tractable algorithm with respect to the number of values appearing in more than two distinct domains. Interestingly, this taxonomy shows that enforcing equality is harder than enforcing difference.

## 1. Introduction

Constraints for reasoning about equality and difference within assignments to a set of variables are ubiquitous in constraint programming. In many settings, one needs to enforce a given degree of diversity or similarity in a solution. For example, in a university timetabling problem we will want to ensure that all courses taken by a particular student are held at different times. Similarly, in meeting scheduling we will want to ensure that the participants of the same meeting are scheduled to meet at the same time and in the same place. Sometimes, when the problem is over-constrained, we might wish to maximise the extent to which these constraints are satisfied. Consider again our timetabling example: we might wish to maximise the number of courses that are scheduled at different times when a student's preferences cannot all be met.

In a constraint programming setting requirements on the diversity and similarity amongst variables can be specified using global constraints. One of the most commonly used global constraints is the AllDifferent (Régin, 1994), which enforces that all variables take pairwise different values. A soft version of the AllDifferent constraint, named SoftAllDiff, has been proposed by Petit, Régin, and Bessiere (2001). They proposed two cost metrics for measuring the degree of satisfaction of the constraint, which are to be minimised or maximised: *graph-* and *variable*-based cost. These two cost metrics are generic and widely used (e.g., van Hoeve, 2004). The former counts the number of equalities, whilst the latter counts the number of variables to change in order to satisfy the corresponding hard constraint. When we wish to enforce that a set of variables take *equal* values, we can use the AllEqual, or its soft variant for the graph-based cost, the SoftAllEqual constraint (Hebrard, O'Sullivan, & Razgon, 2008), or its soft variant for the variable-based cost, the AtMostNValue constraint (Beldiceanu, 2001).

When considering these two constraints (AllDifferent and AllEqual), these two costs (graph-based and variable-based) and objectives (minimisation and maximisation) we can define eight algorithmic problems related to constraints of difference and equality. In fact, because the graph-based costs of AllDifferent and AllEqual are dual, only six distinct problems are thus defined. The structure of this class of constraints is illustrated in Figure 1. For each one, we give the complexity of the best known algorithm for achieving AC and BC. Three of these problems were studied in the past: minimising the cost of SoftAllDiff variable (Petit et al., 2001) and graph-based cost (van Hoeve, 2004) is polynomial whilst maximising the variable-based cost of SoftAllDiff is NP-hard (Bessiere, Hebrard, Hnich, Kiziltan, & Walsh, 2006) for AC and polynomial (Beldiceanu, 2001) for BC. A fourth one, maximising the variable-based cost of the SoftAllEqual constraint, can directly be mapped to a known problem: the Global Cardinality constraint. In this paper,[1] we introduce two efficient algorithms for achieving, respectively, Arc consistency (AC) and Bounds consistency (BC) on the fifth case, minimising the variable-based cost for SoftAllEqual. Moreover, the computational complexity of the last remaining case, maximising the graph-based cost for SoftAllDiff (or, equivalently, minimising the graph-based cost for SoftAllEqual) was still unknown. Informally, this problem is to maximise the number of pairs of variables assigned to a common value. It turns out to be a challenging and interesting problem, in that it is hard but yet can be addressed in several ways. In particular, we show that:

- Finding a solution with at least $k$ pairs of equal variables is NP-complete, hence achieving AC on the corresponding constraint is NP-hard.

- When domains are contiguous, it can be solved in a polynomial number of steps through dynamic programming, hence achieving BC on the corresponding constraint is polynomial.

- There exists a linear approximation by a factor of $\frac{1}{2}$ for the general case.

---

1. Part of the material presented in this paper is based on two conference publications (Hebrard et al., 2008; Hebrard, Marx, O'Sullivan, & Razgon, 2009).

- If no value appears in the domains of more than two distinct variables, then the problem can be solved by a general matching, thus defining another tractable class.

- There exists a fixed parameter tractable algorithm for this problem for a parameter $k$ equal to the number of values that appear in more than two distinct domains.

Moreover, we show that the constraint defined by setting a lower bound on the graph-based cost of SoftAllEqual can be used to efficiently find a set of similar solutions to a set of problems, for instance to promote stability or regularity. Similarly, the dual constraint (SoftAllDiff) can be used to find a set of diverse solutions, for instance to sample a set of configurations. Notice that these two applications have motivated, in part, our choice of cost metrics.

The remainder of this paper is organised as follows. In Section 2 we introduce the necessary technical background. A complete taxonomy of constraints of equality and difference, based on results by other authors as well as original material is presented in Section 3. Then, in the following sections, we present the new results allowing us to close the gaps in this taxonomy. First, in Section 4 we present two efficient algorithm for achieving AC and BC when minimising the variable-based cost of SoftAllEqual. Second, in Section 5 we give a proof of NP-hardness for the problem of achieving AC when maximising the graph-based cost of SoftAllDiff. Third, in Section 6 we present a polynomial algorithm to achieve BC on the same constraint. Finally, in the remaining sections, we explore the algorithmic properties of this preference cost. In Section 7, we show that a natural greedy algorithm approximates the maximum number of equalities within a factor of $\frac{1}{2}$, and that its complexity can be brought down to linear time. Next, in Section 8, we identify a polynomial class for this constraint. Then, in Section 9, we identify a parameter based on this class and show that the SoftAllEqual$_G$ constraint is fixed-parameter tractable with respect to this parameter. Finally, in Section 10, we show how the results obtained in this paper can be applied to sample solutions or, conversely, to promote stability. In particular, we describe two constructions using SoftAllDiff$_G^{min}$ and SoftAllEqual$_G^{min}$ respectively. Concluding remarks are made in Section 11.

## 2. Background

In this section we present the necessary background required by the reader and introduce the notation we use throughout the paper.

### 2.1 Constraint Satisfaction

A constraint satisfaction problem (CSP) is a triplet $\mathcal{P} = (\mathcal{X}, \mathcal{D}, \mathcal{C})$ where $\mathcal{X}$ is a set of variables, $\mathcal{D}$ is a mapping of variables to finite sets of values and $\mathcal{C}$ is a set of constraints that specify allowed combinations of values for subsets of variables. Without loss of generality, we assume $\mathcal{D}(X) \subset \mathbb{Z}$ for all $X \in \mathcal{X}$, and we denote by $\min(X)$ and $\max(X)$ the minimum and maximum values in $\mathcal{D}(X)$, respectively. An assignment of a set of variables $\mathcal{X}$ is a set of pairs $S$ such that $|\mathcal{X}| = |S|$ and for each $X \in \mathcal{X}$, there exists $(X, v) \in S$ with $v \in \mathcal{D}(X)$. A constraint $C \in \mathcal{C}$ is *arc consistent* (AC) iff, when a variable in the scope of $C$ is assigned any value, there exists an assignment to the other variables in $C$ such that $C$ is satisfied. This satisfying assignment is called a *domain support* for the value. Similarly, we call a *range support* an assignment satisfying $C$, but where values, instead of being taken from the domain of each variable ($v \in \mathcal{D}(X)$), can be any integer between the minimum and maximum of this domain following the natural order on $\mathbb{Z}$ ($v \in [\min(X), \ldots, \max(X)]$) . A constraint $C \in \mathcal{C}$ is *range consistent* (RC) iff every value of every variable in the scope of $C$ has a range support. A constraint $C \in \mathcal{C}$ is *bounds consistent* (BC) iff for every variable $X$ in the scope of $C$, $\min(X)$ and $\max(X)$ have a range support. Given a CSP $\mathcal{P} = (\mathcal{X}, \mathcal{D}, \mathcal{C})$, we shall use the following notation throughout the paper: $n$ shall denote the number of variables, i.e., $n = |\mathcal{X}|$; $m$ shall denote the number of distinct unary assignments, i.e., $m = \sum_{X \in \mathcal{X}} |\mathcal{D}(X)|$; $\Lambda$ shall denote the total set of values, i.e., $\Lambda = \bigcup_{X \in \mathcal{X}} \mathcal{D}(X)$; finally, $\lambda$ shall denote the total number of distinct values, i.e., $\lambda = |\Lambda|$.

## 2.2 Soft Global Constraints

Adding a cost variable to a constraint to represent its degree of violation is now common practice in constraint programming. This model was introduced by Petit, Régin, and Bessière (2000). It offers the advantage of unifying hard and soft constraints since arc consistency, along with other types of consistencies, can be applied to such constraints with no extra effort. As a consequence, classical constraint solvers can model over-constrained problems in this way without modification. This approach was applied to a number of other constraints, for instance by van Hoeve, Pesant, and Rousseau (2006). Several cost metrics have been explored for the AllDifferent constraint, as well as several others (e.g., Beldiceanu & Petit, 2004). It is important, if one uses such a unifying model, that the cost metric chosen can be evaluated in polynomial time given a complete assignment of the variables that are constrained. This is the case for the two metrics considered in this paper for the constraints AllDifferent and AllEqual.

The *variable-based cost* counts how many variables need to change in order to obtain a valid assignment for the hard constraint. It can be viewed as the smallest Hamming distance with respect to a satisfying assignment. The *graph-based cost* counts how many times a component of a decomposition of the constraint is violated. Typically these components correspond to edges of a decomposition graph, e.g. for an AllDifferent constraint, the decomposition graph is a clique and an edge is violated if and only if both variables connected by this edge share the same value. The following example, still for the AllDifferent constraint, shows two solutions involving four variables $X_1, \ldots, X_4$ each with domain $\{a, b\}$:

$$S_1 = \{(X_1, a), (X_2, b), (X_3, a), (X_4, b)\}.$$

$$S_2 = \{(X_1, a), (X_2, b), (X_3, b), (X_4, b)\}.$$

In both solutions, at least two variables must change (e.g., $X_3$ and $X_4$) to obtain a valid solution. Therefore, the variable-based cost is 2 for $S_1$ and $S_2$. However, in $S_1$ only two edges are violated, $(X_1, X_3)$ and $(X_2, X_4)$, whilst in $S_2$, three edges are violated, $(X_2, X_3)$, $(X_2, X_4)$ and $(X_3, X_4)$. Thus, the graph-based cost of $S_1$ is 2 whereas it is 3 for $S_2$.

## 2.3 Parameterised Complexity

We shall use the notion of parameterised complexity in Section 9. We refer the reader to Niedermeier's (2006) book for a comprehensive introduction. Given a problem $\mathbf{A}$, a parameterised version of **A** is obtained by specifying a parameter of this problem and getting as additional input a non-negative integer $k$ which restricts the value of this parameter. The resulting parameterised problem $\langle \mathbf{A}, k \rangle$ is *fixed-parameter tractable* (FPT) with respect to $k$ if it can be solved in time $f(k) * n^{\mathcal{O}(1)}$, where $f(k)$ is a function depending only on $k$. When the size of the problem is significantly larger than the parameter $k$, a fixed-parameter algorithm essentially has polynomial behaviour. For instance if $f(k) = 2^k$ then, as long as $k$ is bounded by $\log n$, the problem can be solved in polynomial time.

## 3. Taxonomy

In this section we introduce a taxonomy of soft constraints based on ALLDIFFERENT and ALLEQUAL. We consider the eight algorithmic problems related to constraints of difference and equality defined by combining these two constraints, two costs (graph-based and variable-based), and two objectives (minimisation and maximisation). In fact, because the graph-based costs of ALLDIFFERENT and ALLEQUAL are dual, only six different problems are defined. Observe that we consider only costs defined through inequalities, rather than equalities. There are several reasons for doing so. First, reasoning about the lower bound or the upper bound of the cost variable can yield two extremely different problems, and hence different algorithmic solutions. For instance, we shall see that in some cases the problem is tractable in one direction, and NP-hard in the other direction. When reasoning about cost equality, one will often separate the inference procedures relative to the lower bound, upper bound, and intermediate values. Reasoning about lower and upper bounds is sufficient to model an equality although it might hinder domain filtering when intermediate values for the cost are forbidden. We thus cover equalities in a restricted way, albeit arguably reasonable in practice. Indeed, when dealing with costs and objectives, reasoning about inequalities and bounds is more useful in practice than imposing (dis)equalities.

We close the last remaining cases: the complexity of achieving AC and BC SOFTALLEQUAL$_V^{min}$ in Section 4, that of achieving AC on SOFTALLEQUAL$_G^{min}$ in Section 5 and that of achieving BC on SOFTALLEQUAL$_G^{min}$ in Section 6. Based on these results, Figure 1 can now be completed (fourth and fifth columns).

The next six paragraphs correspond to the six columns of Figure 1, that is, to the twelve elements of the taxonomy. For each of them, we briefly outline the current state of the art, using the following assignment as a recurring example to illustrate the various costs:

$$S_3 = \{(X_1, a), (X_2, a), (X_3, a), (X_4, a), (X_5, b), (X_6, b), (X_7, c)\}.$$

### 3.1 SOFTALLDIFF: **Variable-based cost, Minimisation**

**Definition 1** (SOFTALLDIFF$_V^{min}$)

SOFTALLDIFF$_V^{min}(\{X_1, \ldots, X_n\}, N) \Leftrightarrow N \geq n - |\{v \mid \exists X_i = v\}|$.

Here the cost to minimise is the number of variables that need to be changed in order to obtain a solution satisfying an ALLDIFFERENT constraint. For instance, the cost of $S_3$ is 4 since three of the four variables assigned to $a$ as well as one of the variables assigned to $b$ must change. This objective function was first studied by Petit et al. (2001), and an algorithm for achieving AC in $\mathcal{O}(n\sqrt{m})$ was introduced. To the best of our knowledge, no algorithm with better time complexity for the special case of bounds consistency has been proposed for this constraint. Notice however that Mehlhorn and Thiel's (2000) algorithm achieves BC on the AllDifferent constraint with an $\mathcal{O}(n \log n)$ time complexity. The question of whether this algorithm could be adapted to achieve BC on SoftAllDiff$_V^{min}$ remains open.

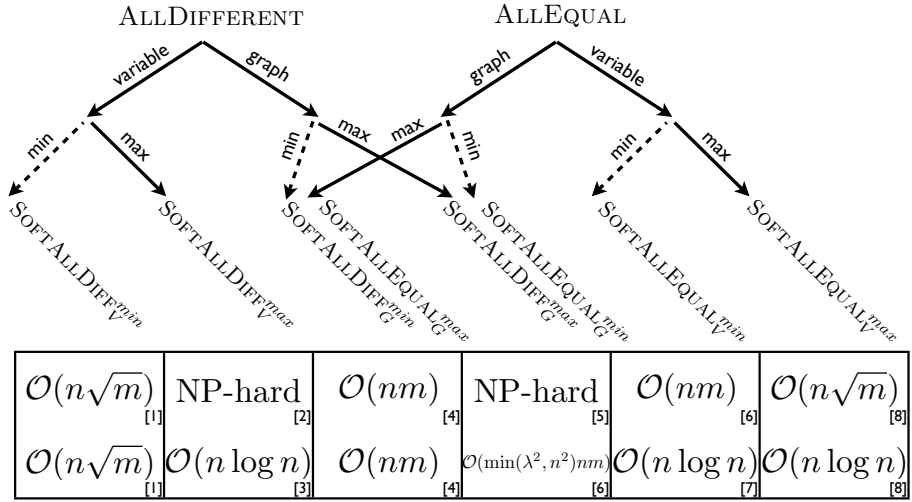

Figure 1: Complexity of optimising difference and equality – first row: AC, second row: BC. Parameter $n$ denotes the number of variables, $m$ the sum of the domain sizes and $\lambda$ the number of distinct values. References: [1] (Petit et al., 2001), [2] (Bessiere et al., 2006), [3] (Beldiceanu, 2001), [4] (van Hoeve, 2004), [5] (Hebrard et al., 2008), [6] (Hebrard et al., 2009), [7] (present paper), [8] (Quimper et al., 2004).

## 3.2 SoftAllDiff: Variable-based cost, Maximisation

**Definition 2** (SoftAllDiff$_V^{max}$)

SoftAllDiff$_V^{max}(\{X_1, \ldots, X_n\}, N) \Leftrightarrow N \le n - |\{v \mid \exists X_i = v\}|$.

Here the same cost is to be maximised. In other words, we want to minimise the number of distinct values assigned to the given set of variables, since the complement of this number to $n$ is exactly the number of variables to modify in order to obtain a solution satisfying an AllDifferent constraint. For instance, the cost of $S_3$ is 4 and the number of distinct values is $7 - 4 = 3$. This constraint was studied under the name AtMostNValue. An algorithm in $\mathcal{O}(n \log n)$ to achieve BC was proposed by Beldiceanu (2001), and a proof that achieving AC is NP-hard was given by Bessiere et al. (2006).

### 3.3 SOFTALLDIFF: **Graph-based cost, Minimisation &** SOFTALLEQUAL: **Graph-based cost, Maximisation**

**Definition 3** (SOFTALLDIFF$_G^{min}$ ≡ SOFTALLEQUAL$_G^{max}$)

$$\text{SOFTALLDIFF}_G^{min}(\{X_1, \ldots, X_n\}, N) \Leftrightarrow N \geq |\{\{i, j\} \mid X_i = X_j \ \& \ i < j\}|.$$

Here the cost to minimise is the number of violated constraints when decomposing ALLDIFFERENT into a clique of binary NOTEQUAL constraints. For instance, the cost of $S_3$ is 7 since four variables share the value $a$ (six violations) and two share the value $b$ (one violation). Clearly, it is equivalent to maximising the number of violated binary EQUAL constraints in a decomposition of a global ALLEQUAL. Indeed, these two costs are complementary to $\binom{n}{2}$ of each other (on $S_3$: $7 + 14 = 21$). An algorithm in $\mathcal{O}(nm)$ for achieving AC on this constraint was introduced by van Hoeve (2004). Again, to our knowledge there is no algorithm improving this complexity for the special case of BC.

### 3.4 SOFTALLEQUAL: **Graph-based cost, Minimisation &** SOFTALLDIFF: **Graph-based cost Maximisation**

**Definition 4** (SOFTALLEQUAL$_G^{min}$ ≡ SOFTALLDIFF$_G^{max}$)

$$\text{SOFTALLEQUAL}_G^{min}(\{X_1, \ldots, X_n\}, N) \Leftrightarrow N \geq |\{\{i, j\} \mid X_i \neq X_j \ \& \ i < j\}|.$$

Here we consider the same two complementary costs, however we aim at optimising in the opposite way. In Section 5 we show that achieving AC on this constraint is NP-hard and, in Section 6 we show that, when domains are contiguous intervals, computing the optimal cost can be done in $\mathcal{O}(\min(n\lambda^2, n^3))$. As a consequence, BC can be achieved in polynomial time.

### 3.5 SOFTALLEQUAL: **Variable-based cost, Minimisation**

**Definition 5** (SOFTALLEQUAL$_V^{min}$)

$$\text{SOFTALLEQUAL}_V^{min}(\{X_1, \ldots, X_n\}, N) \Leftrightarrow N \geq n - \max_{v \in \Lambda}(|\{i \mid X_i = v\}|).$$

Here the cost to minimise is the number of variables that need to be changed in order to obtain a solution satisfying an ALLEQUAL constraint. For instance, the cost of $S_3$ is 3 since four variables already share the same value. This is equivalent to maximising the number of variables sharing a given value. Therefore this bound can be computed trivially by counting the occurrences of every value in the domains. However, pruning the domains according to this bound without degrading the time complexity is not as trivial. In Section 4, we introduce two filtering algorithms, achieving AC and RC in the same complexity as that of counting values.

### 3.6 SOFTALLEQUAL: **Variable-based cost, Maximisation**

**Definition 6** (SOFTALLEQUAL$_V^{max}$)

$$\text{SOFTALLEQUAL}_V^{max}(\{X_1, \ldots, X_n\}, N) \Leftrightarrow N \leq n - \max_{v \in \Lambda}(|\{i \mid X_i = v\}|).$$

Here the same cost has to be maximised. In other words we want to minimise the maximum cardinality of each value. For instance, the cost of $S_3$ is 3, that is, the complement to $n$ of the maximum cardinality of a value ($3 = 7 - 4$). This is exactly equivalent to applying a GLOBAL CARDINALITY constraint (considering only the upper bounds on the cardinalities). Two algorithms, for achieving AC and BC on this constraint and running in $\mathcal{O}(\sqrt{n}m)$ and $\mathcal{O}(n \log n)$ respectively, was introduced by Quimper et al. (2004).

## 4. The Complexity of Arc and Bounds Consistency on SOFTALLEQUAL$_V^{min}$

Here we show how to achieve AC, RC and BC on the SOFTALLEQUAL$_V^{min}$ constraints (see Definition 5). This constraint is satisfied if and only if $n$ minus the cardinality of any set of variables assigned to a single value is less than or equal to the value of the cost variable $N$. In other words, it is satisfied if there are at least $k$ variables sharing a value, where $k = n - max(N)$. Therefore, for simplicity sake, we shall consider the following equivalent formulation, where $N$ is a lower bound on the complement to $n$ of the same cost ($N' = n - N$):

$$N' \leq \max_{v \in \Lambda}(|\{i \mid X_i = v\}|).$$

We shall see that to filter the domain of $N'$ and the $X_i$'s we need to compute two properties:

1. An upper bound $k^*$ on the number of occurrences amongst all values.

2. The set of values that can actually appear $k^*$ times.

Computing the set of values that appear in the largest possible number of variable domains can be performed trivially in $\mathcal{O}(m)$, by counting the number of occurrences of every value, i.e., the number of variables whose domain contains $v$.

However, if domains are discrete intervals defined by lower and upper bounds, it can be done even more efficiently. Given two integers $a$ and $b$, $a \leq b$, we say that the set of all integers $x$, $a \leq x \leq b$, is an *interval* and denote it by $[a, b]$. In the rest of this section we shall assume that the overall set of values values $\Lambda = \bigcup_{X \in \mathcal{X}} \mathcal{D}(X)$ is the interval $[1, \lambda]$.

**Definition 7 (Occurrence function and derivative)** *Given a constraint network $\mathcal{P} = (\mathcal{X}, \mathcal{D}, \mathcal{C})$, the occurrence function occ is the mapping from values in $\Lambda$ to $\mathbb{N}$ defined as follows:*

$$occ(v) = |\{X \mid X \in \mathcal{X} \;\&\; v \in doms(X)\}|.$$

*The "derivative" of occ, $\delta_{occ}$, maps each value $v \in \Lambda$ to the difference between the value of $occ(v-1)$ and $occ(v)$:*

$$\begin{aligned} \delta_{occ}(0) &= 0, \\ \delta_{occ}(v) &= occ(v) - occ(v-1). \end{aligned}$$

We give an example of the occurrence function for a set of variables with interval domains in Figure 2.

Algorithm 1 computes $occ^{-1}$, that is, the inverse of the occurrence function, which maps every element in the interval $[1, n]$ to the set of values appearing that many times. It runs

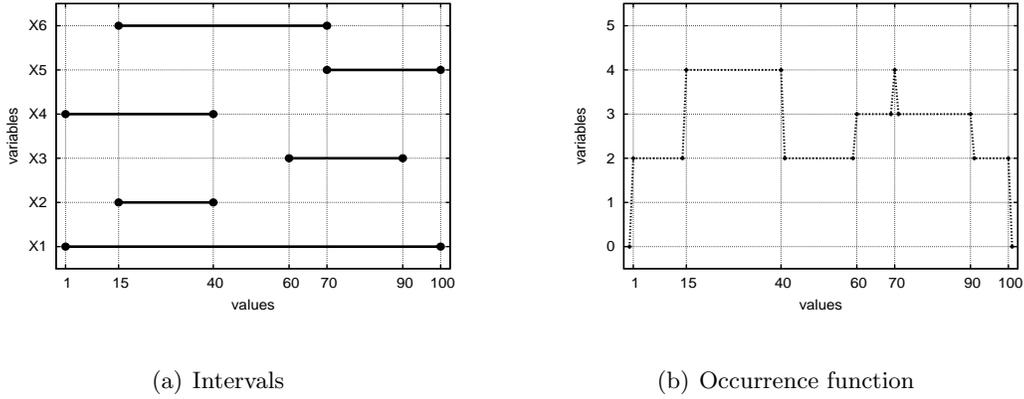

(a) Intervals              (b) Occurrence function

Figure 2: A set of intervals (a) and the corresponding occurrence function (b).

---

**Algorithm 1:** Computing the inverse occurrence function.

**Data**: A set of variables: $\mathcal{X}$
**Result**: $occ^{-1} : [1, n] \mapsto 2^{\Lambda}$

$\delta_{occ}(v) \leftarrow \emptyset$;

**1 foreach** $X \in \mathcal{X}$ **do**
    | $\delta_{occ}(\min(X)) \leftarrow \delta_{occ}(\min(X)) + 1$;
    | $\delta_{occ}(\max(X) + 1) \leftarrow \delta_{occ}(\max(X) + 1) - 1$;

**2** $\forall x \in [1, n], \; occ^{-1}(x) \leftarrow \emptyset$;

$x \leftarrow 0$;

pop first element $(v, a)$ of $\delta_{occ}$;

**repeat**
    | pop first element $(w, b)$ of $\delta_{occ}$;
    | $x \leftarrow x + \delta_{occ}(a)$;
    | $occ^{-1}(x) \leftarrow occ^{-1}(x) \cup [a, b - 1]$;
    | $a \leftarrow b$;

**until** $\delta_{occ} = \emptyset$;

---

in $\mathcal{O}(n \log n)$ worst-case time complexity if we assume it is easy to extract both an upper bound $(k^* \geq N')$ and the set of values that can appear $k^*$ times from $occ^{-1}$.

The idea behind this algorithm, which we shall reuse throughout this paper, is that when domains are given as discrete intervals one can compute the non-null values of the derivative $\delta_{occ}$ of the occurrence function $occ$ in $\mathcal{O}(n \log n)$ time. The procedure is closely related to the concept of *sweep* algorithms (Beldiceanu & Carlsson, 2001) used, for instance, to implement filtering algorithms for the CUMULATIVE constraint. Instead of scanning the entire horizon, one can jump from an event to the next, assuming that nothing changes between two events. As in the case of the CUMULATIVE constraint, events here correspond to start and end points of the domains. In fact, it is possible to compute the same lower bound, with the same complexity, by using Petit, Régin, and Bessiere's (2002) Range-based Max-CSP Algorithm (RMA)[2] on a reformulation as a Max-CSP. Given a set of variables $\mathcal{X}$, we add an extra variable $Z$ whose domain is the union of all domains in $\mathcal{X}$: $\mathcal{D}(Z) = \Lambda = \bigcup_{X \in \mathcal{X}} \mathcal{D}(X)$. Then

---

2. We thank the anonymous reviewer who made this observation.

we link it to other variables in $\mathcal{X}$ through binary equality constraints:

$$\forall X \in \mathcal{X}, \ Z = X.$$

There is a one-to-one mapping between the solutions of this Max-CSP and the satisfying assignments of a SoftAllEqual$_V^{min}$ constraint on $(\mathcal{X}, N)$, where the value of $N$ corresponds to the number of violated constraints in the Max-CSP. The lower bound on the number of violations computed by RMA and the lower bound $k^*$ on $N$ computed in Algorithm 1 are, therefore, the same. Moreover the procedures are essentially equivalent, i.e., modulo the modelling step. Algorithm 1 can be seen as a particular case of RMA: the same ordered set of intervals is computed, and subsequently associated with a violation cost. However, we use our formalism, since the notion of occurrence function and its derivative is important and used throughout the paper.

We first define a simple data structure that we shall use to compute and represent the function $\delta_{occ}$. A specific data structure is required since indexing the image of $\delta_{occ}(v)$ by the value $v$ would add a factor of $\lambda$ to the (space and therefore time) complexity. The non-zero values of $\delta_{occ}$ are stored as a list of pairs whose first element is a value $v \in [1, \ldots, \lambda]$ and second element stands for $\delta_{occ}(v)$. The list is maintained in increasing order of the pair's first element. Given an ordered list $\delta_{occ} = [(v_1, o_1), \ldots, (v_k, o_k)]$, the assignment operation $\delta_{occ}(v_i) \leftarrow o_i$ can therefore been done in $\mathcal{O}(\log |\delta_{occ}|)$ steps as follows:

1. The rank $r$ of the pair $(v_j, o_j)$ such that $v_j$ is minimum and $v_j \geq v_i$ is computed through a dichotomic search.

2. If $v_i = v_j$, the pair $(v_j, o_j)$ is removed.

3. The pair $(v_i, o_i)$ is inserted at rank $r$.

Moreover, one can access the element with minimum (resp. maximum) first element in constant time since it is first (resp. last) in the list. Finally, the value of $\delta_{occ}(v_i)$ is $o_i$ if there exists a pair $(v_j, o_j)$ in the list, and 0 otherwise. Computing this value can also be done in logarithmic time.

The derivative $\delta_{occ}(v)$ is computed in Loop 1 of Algorithm 1 using the assignment operator defined above. Observe that if $\mathcal{D}(X) = [a, b]$, then $X$ contributes only to two values of $\delta_{occ}$: it increases $\delta_{occ}(a)$ by 1 and decreases $\delta_{occ}(b + 1)$ by 1. For every value $w$ such that there is no $X$ with $\min(X) = w$ or $\max(X) + 1 = w$, $\delta_{occ}(w)$ is null. In other words, we can define $\delta_{occ}(v)$ for any value $v$, as follows:

$$\delta_{occ}(v) = (|\{i \mid \min(X_i) = v\}| - |\{i \mid \max(X_i) = v - 1\}|).$$

Therefore, by going through every variable $X \in \mathcal{X}$, we can compute the non-null values of $\delta_{occ}$ in time $\mathcal{O}(n \log n)$ using the simple list structure described above.

Then, starting from Line 2, we compute $occ^{-1}$ by going through the non-zero values $v$ of the derivative, i.e. such that $\delta_{occ}(v) \neq 0$, in increasing order of $v$. Recall that we use an ordered list, so this is trivially done in linear time. By definition, the occurrence function is constant on the interval defined by two such successive values. Since the number of non-zero values of $\delta_{occ}$ is bounded by $\mathcal{O}(n)$, the overall worst-case time complexity is in $\mathcal{O}(n \log n)$. We use Figure 3 (a,c & d) to illustrate an execution of Algorithm 1. First, six variables and

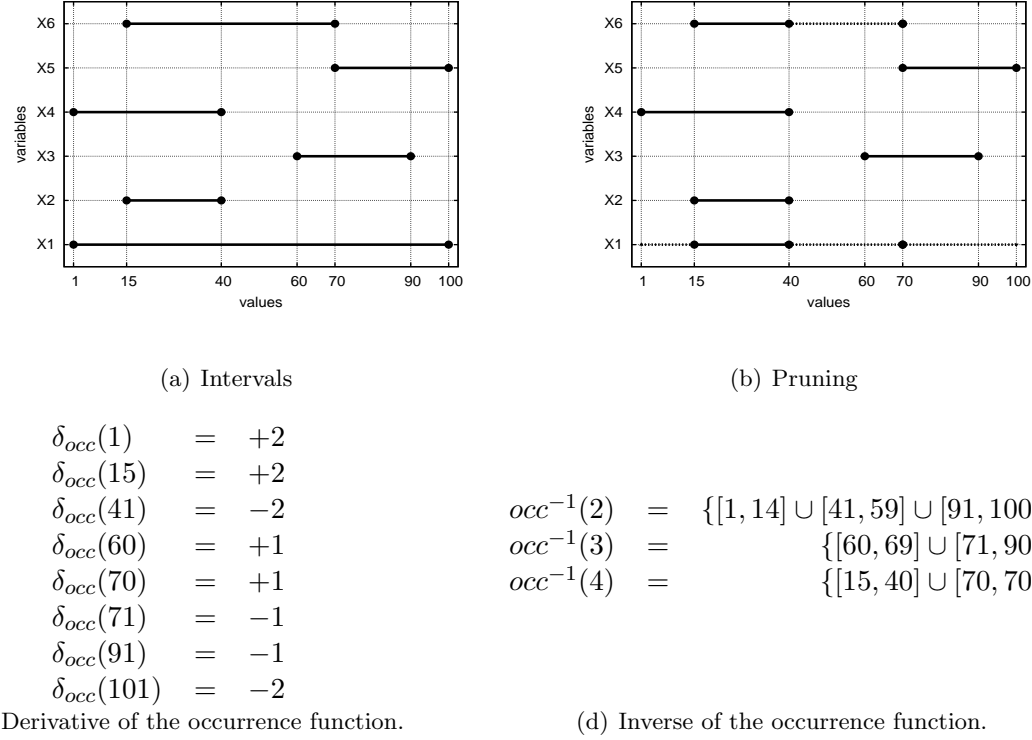

(a) Intervals

(b) Pruning

$$\begin{aligned}
\delta_{occ}(1) &= +2 \\
\delta_{occ}(15) &= +2 \\
\delta_{occ}(41) &= -2 \\
\delta_{occ}(60) &= +1 \\
\delta_{occ}(70) &= +1 \\
\delta_{occ}(71) &= -1 \\
\delta_{occ}(91) &= -1 \\
\delta_{occ}(101) &= -2
\end{aligned}$$

$$\begin{aligned}
occ^{-1}(2) &= \{[1,14] \cup [41,59] \cup [91,100]\} \\
occ^{-1}(3) &= \{[60,69] \cup [71,90]\} \\
occ^{-1}(4) &= \{[15,40] \cup [70,70]\}
\end{aligned}$$

(c) Derivative of the occurrence function.

(d) Inverse of the occurrence function.

Figure 3: Execution of Algorithm 1: A set of intervals (a). The same set of intervals where inconsistent sub-intervals for a lower bound on the number of equalities of 4 (N' ≥ 4) are represented as dashed lines (b). (c) and (d) represent the derivative, and the inverse of the occurrence function for the initial set of intervals, respectively.

their domains are represented in Figure 3(a). Then, in Figures 3(c) and 3(d) we show the derivative and the inverse, respectively, of the occurrence function.

Alternatively, when $\lambda < n \log n$, it is possible to compute $occ^{-1}$ in $\mathcal{O}(n+\lambda)$ by replacing the data structure used to store $\delta_{occ}$ by a simple array, indexed by values in $[1, \lambda]$. Accessing and updating a value of $\delta_{occ}$ can thus be done in constant time.

Now we show how to prune the variables in $\mathcal{X}$ with respect to this bound without degrading the time complexity. According to the method used we can, therefore, achieve AC or RC in a worst-case time complexity of $\mathcal{O}(m)$ or $\mathcal{O}(\min(n+\lambda, n \log n)$, respectively.

**Theorem 1** *Enforcing* AC *(resp.* RC*) on* SoftAllEqual$_V^{min}$ *can be achieved in in* $\mathcal{O}(m)$ *steps (resp.* $\mathcal{O}(\min(n+\lambda, n \log n))$*).*

**Proof.** We suppose, without loss of generality, that the current lower bound on $N'$ is $k$. We first compute the inverse occurrence function either by counting values, or considering interval domains using Algorithm 1. From this we can define the set of values with highest number of occurrences. Let this number of occurrences be $k^*$, and the corresponding set of values be $V$ (i.e. $occ^{-1}(k^*) = V$). Then there are three cases to consider:

1. First, if every value appears in strictly fewer than $k$ domains ($k^* < k$) then the constraint is violated.

2. Second, if at least one value $v$ appears in the domains of at least $k + 1$ variables ($k^* > k$), then we can build a support for every value $w \in \mathcal{D}(X)$. Let $v \in V$, we assign all the variables in $\mathcal{X} \setminus X$ with $v$ when possible. The resulting assignment has at least $k$ occurrences of $v$, hence it is consistent. Consequently, since $k^* > k$, every value is consistent.

3. Otherwise, if neither of the two cases above hold, we know that no value appears in more than $k$ domains, and that at least one appears $k$ times. Recall that $V$ denotes the set of such values. In this case, the pair $(X, v)$ is inconsistent if and only if $v \notin V$ & $V \subset \mathcal{D}(X)$.

   We first suppose that this condition does not hold and show that we can build a support. If $v \in V$ then clearly we can assign every possible variable to $v$ and achieve a cost of $k$. If $V \not\subset \mathcal{D}(X)$, then we consider $w$ such that $w \in V$ and $w \notin \mathcal{D}(X)$. By assigning every variable with $w$ when possible we achieve a cost of $k$ no matter what value is assigned to $X$.

   Now we suppose that $v \notin V$ & $V \subset \mathcal{D}(X)$ holds and show that $(X, v)$ does not have an AC support. Indeed, once $X$ is assigned to $v$ the domains are such that no value appears in $k$ domains or more, since *every* value in $V$ has now one fewer occurrence, hence we are back to Case 1.

Computing the set $V$ of values satisfying the condition above can be done easily once the inverse occurrence function has been computed. On the one hand, if this function $occ^{-1}$ has been computed by counting every value in every domain, then the supports used in the proofs are all domain supports, hence AC is achieved. On the other hand, if domains are approximated by their bounds and Algorithm 1 is used instead, the supports are all range supports, hence RC is achieved. In Case 3, the domain can be pruned down to the set $V$ of values whose number of occurrences is $k$, as illustrated in Figure 3 (b). □

**Corollary 1** *Enforcing* BC *on* SOFTALLEQUAL$_V^{min}$ *can be achieved in* $\mathcal{O}(\min(n + \lambda, n \log n))$ *steps.*

**Proof.** This is a direct implication of Theorem 1. □

The proof of Theorem 1 yields a domain filtering procedure. Algorithm 2 achieves either AC or RC depending on the version of Algorithm 1 used in Line 1 to compute the inverse occurrence function. The later function $occ^{-1}$ is then used in Line 2, 3 and 4 to, respectively, catch a global inconsistency, prune the upper bound of $N'$ and prune the domains of the variables in $\mathcal{X}$.

Figure 3(b) illustrates the pruning that one can achieve on $\mathcal{X}$ provided that the lower bound on $N'$ is equal to 4. Dashed lines represent inconsistent intervals. The set $V$ of values used in Line 4 of Algorithm 2 is $occ^{-1}(4) = \{[15, 40] \cup [70, 70]\}$.

---

**Algorithm 2:** Propagation of $\textsc{SoftAllEqual}_V^{min}(\{X_1, \ldots, X_n\}, N')$.

---

**1**   $occ^{-1} \leftarrow$ Algorithm 1;

    $ub \leftarrow n$;

    **while** $occ^{-1}(ub) = \emptyset$ **do**

      $\lfloor$   $ub \leftarrow ub - 1$;

**2**   **if** $min(N') > ub$ **then** fail;

    **else**

**3**     |   $max(N') \leftarrow ub$;

    |   **if** $min(N') = max(N')$ **then**

    |   |   $V \leftarrow occ^{-1}(min(N'))$;

**4**     |   |   **foreach** $X \in \mathcal{X}$ **do**   **if** $V \subset \mathcal{D}(X)$ **then**   $\mathcal{D}(X) \leftarrow V$;

---

## 5. The Complexity of Arc Consistency on $\textsc{SoftAllEqual}_G^{min}$

Here we show that achieving AC on $\textsc{SoftAllEqual}_G^{min}$ is NP-hard. In order to achieve AC we need to compute an arc consistent lower bound on the cost variable $N$ constrained as follows:

$$N \leq |\{\{i, j\} \mid X_i \neq X_j \ \& \ i < j\}|.$$

In other words, we want to find an assignment of the variables in $\mathcal{X}$ minimising the number of pairwise disequalities, or maximising the number of pairwise equalities. We consider the corresponding decision problem ($\textsc{SoftAllEqual}_G^{min}$-decision), and show that it is NP-hard through a reduction from **3dMatching** (Garey & Johnson, 1979).

**Definition 8** ($\textsc{SoftAllEqual}_G^{min}$-decision)
Data: *An integer $N$, a set $\mathcal{X}$ of variables.*
Question: *Does there exist a mapping $s : \mathcal{X} \mapsto \Lambda$ such that $\forall X \in \mathcal{X}, \ s[X] \in \mathcal{D}(X)$ and $|\{\{i, j\} \mid s[X_i] = s[X_j] \ \& \ i \neq j\}| \geq N$?*

**Definition 9 (3dMatching)**
Data: *An integer $K$, three disjoint sets $X, Y, Z$, and $T \subseteq X \times Y \times Z$.*
Question: *Does there exist $M \subseteq T$ such that $|M| \geq K$ and $\forall m_1, m_2 \in M, \forall i \in \{1, 2, 3\}, \ m_1[i] \neq m_2[i]$?*

**Theorem 2 (The Complexity of $\textsc{SoftAllEqual}_G^{min}$)** *Finding a satisfying assignment for the $\textsc{SoftAllEqual}_G^{min}$ constraint is NP-complete even if no value appears in more than three domains.*

**Proof.** The problem $\textsc{SoftAllEqual}_G^{min}$-decision is clearly in NP: checking the number of equalities in an assignment can be done in $\mathcal{O}(n^2)$ time.

We use a reduction from **3dMatching** to show completeness. Let $P = (X, Y, Z, T, K)$ be an instance of **3dMatching**, where: $K$ is an integer; $X, Y, Z$ are three disjoint sets such that $X \cup Y \cup Z = \{x_1, \ldots, x_n\}$; and $T = \{t_1, \ldots, t_m\}$ is a set of triplets over $X \times Y \times Z$. We build an instance $I$ of $\textsc{SoftAllEqual}_G^{min}$ as follows:

1. Let $n = |X| + |Y| + |Z|$, we build $n$ variables $\{X_1, \ldots, X_n\}$.

2. For each $t_l = \langle x_i, x_j, x_k \rangle \in T$, we have $l \in \mathcal{D}(X_i)$, $l \in \mathcal{D}(X_j)$ and $l \in \mathcal{D}(X_k)$.

3. For each pair $(i, j)$ such that $1 \leq i < j \leq n$, we put the value $(|T| + (i-1) * n + j)$ in both $\mathcal{D}(X_i)$ and $\mathcal{D}(X_j)$.

We show there exists a matching of $P$ of size $K$ if and only if there exists a solution of $I$ with $\lfloor \frac{3K+n}{2} \rfloor$ equalities. We refer to "a matching of $P$" and to a "solution of $I$" as "a matching" and "a solution" throughout this proof, respectively.

$\Rightarrow$: We show that if there exists a matching of cardinality $K$ then there exists a solution with at least $\lfloor \frac{3K+n}{2} \rfloor$ equalities. Let $M$ be a matching of cardinality $K$. We build a solution as follows. For all $t_l = \langle x_i, x_j, x_k \rangle \in M$ we assign $X_i$, $X_j$ and $X_k$ to $l$ (item 2 above). Observe that there remain exactly $n - 3K$ unassigned variables after this process. We pick an arbitrary pair of unassigned variables and assign them with their common value (item 3 above), until at most one variable is left (if one variable is left we assign it to an arbitrary value). Therefore, the solution obtained in this way has exactly $\lfloor \frac{3K+n}{2} \rfloor$ equalities, $3K$ from the variables corresponding to the matching and $\lfloor \frac{n-3K}{2} \rfloor$ for the remaining variables.

$\Leftarrow$: We show that if the cardinality of the maximal matching is $K$, then there is no solution with more than $\lfloor \frac{3K+n}{2} \rfloor$ equalities. Let $S$ be a solution. Furthermore, let $L$ be the number of values appearing three times in $S$. Observe that this set of values corresponds to a matching. Indeed, a value $l$ appears in three domains $\mathcal{D}(X_i), \mathcal{D}(X_j)$ and $\mathcal{D}(X_k)$ if and only if there exists a triplet $t_l = \langle x_i, x_j, x_k \rangle \in T$ (item 2 above). Since a variable can only be assigned to a single value, the values appearing three times in a solution form a matching. Moreover, since no value appears in more than three domains, all other values can appear at most twice. Hence the number of equalities in $S$ is less than or equal to $\lfloor \frac{3L+n}{2} \rfloor$, where $L$ is the size of a matching. It follows that if there is no matching of cardinality greater than $K$, there is no solution with more than $\lfloor \frac{3K+n}{2} \rfloor$ equalities. $\qquad \square$

Cohen, Cooper, Jeavons, and Krokhin (2004) showed that the language of soft binary equality constraints is NP-complete, for as few as three distinct values. On the one hand, Theorem 2 applies to a more specific class of problems where the constraint network formed by the soft binary constraints is a clique. On the other hand, the proof requires an unbounded number of values, these two results are therefore incomparable. However, we shall see in Section 9 that this problem is fixed parameter tractable with respect to the number of values, hence polynomial when it is bounded.

## 6. The Complexity of Bounds Consistency on SoftAllEqual$_G^{min}$

In this section we introduce an efficient algorithm that, assuming the domains are discrete intervals, computes the maximum possible pairs of equal values in an assignment. We therefore need to solve the optimisation version of the problem defined in the previous section (Definition 8):

**Definition 10** (SoftAllEqual$_G^{min}$-optimisation)
Data: *A set $\mathcal{X}$ of variables.*
Question: *What is the maximum integer $K$ such that there exists a mapping $s : \mathcal{X} \mapsto \Lambda$ satisfying $\forall X \in \mathcal{X}$, $s[X] \in \mathcal{D}(X)$ and $|\{\{i, j\} \mid s[X_i] = s[X_j] \ \& \ i \neq j\}| = K$?*

The algorithm we introduce allows us to close the last remaining open complexity question in Figure 1: bc on the SoftAllEqual$_G^{min}$ constraint. We then improve it by reducing the time complexity thanks to a preprocessing step.

We use the same terminology as in Section 4, and refer to the set of all integers $x$ such that $a \leq x \leq b$ as the interval $[a, b]$. Let $\mathcal{X}$ be the set of variables of the considered CSP and assume that the domains of all the variables of $\mathcal{X}$ are sub-intervals of $[1, \lambda]$. We denote by $\mathbf{ME}(\mathcal{X})$ the set of all assignments $P$ to the variables of $\mathcal{X}$ such that the number of pairs of equal values of $P$ is the maximum possible. The subset of $\mathcal{X}$ containing all the variables whose domains are subsets of $[a, b]$ is denoted by $\mathcal{X}_{a,b}$. The subset of $\mathcal{X}_{a,b}$ including all the variables containing the given value $c$ in their domains is denoted by $\mathcal{X}_{a,b,c}$. Finally the number of pairs of equal values in an element of $\mathbf{ME}(\mathcal{X}_{a,b})$ is denoted by $C_{a,b}(\mathcal{X})$ or just $C_{a,b}$ if the considered set of variables is clear from the context. For notational convenience, if $b < a$, then we set $\mathcal{X}_{a,b} = \emptyset$ and $C_{a,b} = 0$. The value $C_{1,\lambda}(\mathcal{X})$ is the number of equal pairs of values in an element of $\mathbf{ME}(\mathcal{X})$.

**Theorem 3** $C_{1,\lambda}(\mathcal{X})$ *can be computed in* $\mathcal{O}((n + \lambda)\lambda^2)$ *steps.*

**Proof.** The problem is solved by a dynamic programming approach: for every $a, b$ such that $1 \leq a \leq b \leq \lambda$, we compute $C_{a,b}$. The main observation that makes it possible to use dynamic programming is the following: in every $P \in \mathbf{ME}(\mathcal{X}_{a,b})$ there is a value $c$ ($a \leq c \leq b$) such that every variable $X \in \mathcal{X}_{a,b,c}$ is assigned value $c$. To see this, let value $c$ be a value that is assigned by $P$ to a maximum number of variables. Suppose that there is a variable $X$ with $c \in \mathcal{D}(X)$ that is assigned by $P$ to a different value, say $c'$. Suppose that $c$ and $c'$ appear on $x$ and $y$ variables, respectively. By changing the value of $X$ from $c'$ to $c$, we increase the number of equalities by $x - (y - 1) \geq 1$ (since $x \geq y$), contradicting the optimality of $P$.

Notice that $\mathcal{X}_{a,b} \setminus \mathcal{X}_{a,b,c}$ is the disjoint union of $\mathcal{X}_{a,c-1}$ and $\mathcal{X}_{c+1,b}$ (if $c - 1 < a$ or $c + 1 > b$, then the corresponding set is empty). These two sets are independent in the sense that there is no value that can appear on variables from both sets. Thus it can be assumed that $P \in \mathbf{ME}(\mathcal{X}_{a,b})$ restricted to $\mathcal{X}_{a,c-1}$ and $\mathcal{X}_{c+1,b}$ are elements of $\mathbf{ME}(\mathcal{X}_{a,c-1})$ and $\mathbf{ME}(\mathcal{X}_{c+1,b})$, respectively. Taking into consideration all possible values $c$, we get

$$C_{a,b} = \max_{c, a \leq c \leq b} \left( \binom{|\mathcal{X}_{a,b,c}|}{2} + C_{a,c-1} + C_{c+1,b} \right). \tag{1}$$

In the first step of Algorithm 3, we compute $|\mathcal{X}_{a,b,c}|$ for all values of $a, b, c$. For each triple $a, b, c$, it is easy to compute $|\mathcal{X}_{a,b,c}|$ in time $\mathcal{O}(n)$, hence all these values can be computed in time $\mathcal{O}(n\lambda^3)$. However, the running time can be reduced to $\mathcal{O}((n + \lambda)\lambda^2)$ by using the same idea as in Algorithm 1. For each pair $a, b$, we compute the number of occurrences of each value $c$ by first computing a derivative $\delta_{a,b}$. More precisely, we define $\delta_{a,b}(c) = |\mathcal{X}_{a,b,c}| - |\mathcal{X}_{a,b,c-1}|$ and compute $\delta_{a,b}(c)$ for every $a < c \leq b$ (Algorithm 3, Line 1-2). Thus by going through all the variables, we can compute the $\delta_{a,b}(c)$ values for a fixed $a, b$ and for all $a \leq c \leq b$ in time $\mathcal{O}(n)$ and we can also compute $|\mathcal{X}_{a,b,a}|$ in the same time bound. Now it is possible to compute the values $|\mathcal{X}_{a,b,c}|$, $a < c \leq b$ in time $\mathcal{O}(\lambda)$ by using the equality $|\mathcal{X}_{a,b,c}| = |\mathcal{X}_{a,b,c-1}| + \delta_{a,b}(c)$ iteratively (Algorithm 3, Line 3).

In the second step of the algorithm, we compute all the values $C_{a,b}$. We compute these values in increasing order of $b - a$. If $a = b$, then $C_{a,b} = \binom{|\mathcal{X}_{a,a,a}|}{2}$. Otherwise, values $C_{a,c-1}$ and $C_{c+1,b}$ are already available for every $a \leq c \leq b$, hence $C_{a,b}$ can be determined in time $\mathcal{O}(\lambda)$ using Eq. (1) (Algorithm 3, Line 4). Thus all the values $C_{a,b}$ can be computed in time

---

**Algorithm 3:** Computing the maximum number of equalities.

**Data**: A set of variables: $\mathcal{X}$
**Result**: $C_{1,\lambda}(\mathcal{X})$
$\forall\ 1 \leq a,b,c \leq \lambda,\ \delta_{a,b}(c) \leftarrow |\mathcal{X}_{a,b,c}| \leftarrow C_{a,b} \leftarrow 0;$
**foreach** $k \in [0, \lambda - 1]$ **do**
$\quad$ **foreach** $a \in [1, \lambda - k]$ **do**
$\quad\quad b \leftarrow a + k;$
$\quad\quad$ **foreach** $X \in \mathcal{X}_{a,b}$ **do**
**1** $\quad\quad\quad \delta_{a,b}(\min(X)) \leftarrow \delta_{a,b}(\min(X)) + 1;$
**2** $\quad\quad\quad \delta_{a,b}(\max(X) + 1) \leftarrow \delta_{a,b}(\max(X) + 1) - 1;$
$\quad\quad$ **foreach** $c \in [a, b]$ **do**
**3** $\quad\quad\quad |\mathcal{X}_{a,b,c}| \leftarrow |\mathcal{X}_{a,b,c-1}| + \delta_{a,b}(c);$
**4** $\quad\quad\quad C_{a,b} \leftarrow \max(C_{a,b}, (\binom{|\mathcal{X}_{a,b,c}|}{2} + C_{a,c-1} + C_{c+1,b}));$

$\quad$ return $C_{1,\lambda};$

---

$\mathcal{O}(\lambda^3)$, including $C_{1,\lambda}$, which is the value of the optimum solution of the problem. Using standard techniques (storing for each $C_{a,b}$ a value $c$ that minimises (1)), a third step of the algorithm can actually produce a variable assignment that obtains the maximum value. $\quad\square$

| $C_{a,b}$ | $a = 1$ | $a = 2$ | $a = 3$ | $a = 4$ |
|---|---|---|---|---|
| $b = 1$ | 1 | | | |
| $b = 2$ | $\mathcal{X}_{1,2,1} + C_{2,2} = 3$ | 0 | | |
| $b = 3$ | $\mathcal{X}_{1,3,1} + C_{2,3} = 6$ | 0 | 0 | |
| $b = 4$ | $\mathcal{X}_{1,4,1} + C_{2,4} = 16$ | $\mathcal{X}_{2,4,4} + C_{2,3} = 6$ | $\mathcal{X}_{3,4,4} + C_{3,3} = 3$ | 1 |

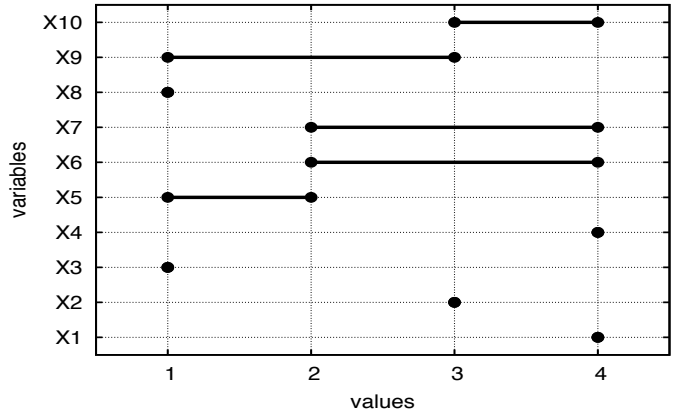

Figure 4: A set of intervals, and the corresponding dynamic programming Table ($C_{a,b}$).

Algorithm 3 computes the largest number of equalities one can achieve by assigning a set of variables with interval domains. It can therefore be used to find an optimal solution to either SoftAllDiff$_G^{max}$ or SoftAllEqual$_G^{min}$. Notice that for the latter one needs to take the complement to $\binom{n}{2}$ in order to get the value of the violation cost. Clearly, it follows that achieving range or bounds consistency on these two constraints can be done in polynomial time, since Algorithm 3 can be used as an oracle for testing the existence of a range support. We give an example of the execution of Algorithm 3 in Figure 4. A set of ten variables, from $X_1$ to $X_{10}$ are represented. Then we give the the table $C_{a,b}$ for all pairs $a, b \in [1, \lambda]$.

The complexity can be further reduced if $\lambda \gg n$. Here again, we will use the occurrence function, albeit in a slightly different way. The intuition is that some values and intervals of values are dominated by other. When the occurrence function is monotonically increasing, it means that we are moving toward dominating values (they can be taken by a larger set of variables), and conversely, a monotonic decrease denotes dominated values. Notice that since we are considering discrete values, some variations may not be apparent in the occurrence function. For instance, consider two variables $X$ and $Y$ with respective domains $[a, b]$ and $[b + 1, c]$ such that $a \leq b \leq c$. The occurrence function for these two variables is constant on $[a, c]$. However, for our purpose, we need to distinguish between "true" monotonicity and that induced by the discrete nature of the problem. We therefore consider some rational values when defining the occurrence function. In the example above, by introducing an extra point $b + \frac{1}{2}$ to the occurrence function, we can now capture the fact that in fact it is not monotonic on $[a, c]$.

Let $\mathcal{X}$ be a set of variables with interval domains in $[1, \lambda]$. Consider the occurrence function $occ : Q \mapsto [0..n]$, where $Q \subset \mathbb{Q}$ is a set of values of the form $a/2$ for some $a \in \mathbb{N}$, such that $\min(Q) = 1$ and $\max(Q) = \lambda$. Intuitively, the value of $occ(a)$ is the number of variables whose domain interval encloses the value $a$, more formally:

$$\forall a \in Q, \; occ(a) = |\{X \mid X \in \mathcal{X}, \min(X) \leq a \leq \max(X)\}|.$$

Such a function, along with the corresponding set of intervals, is depicted in Figure 5. A *crest* of the function $occ$ is an interval $[a, b] \subseteq Q$ such that for some $c \in [a, b]$, $occ$ is monotonically increasing on $[a, c]$ and monotonically decreasing on $[c, b]$. For instance, on the set intervals represented in Figure 5, $[1, 15]$ is a crest since it is monotonically increasing on $[1, 12]$ and monotonically decreasing on $[12, 15]$.

Let $\mathcal{I}$ be a partition of $[1, \lambda]$ into a set of intervals such that every element of $\mathcal{I}$ is a crest. For instance, $\mathcal{I} = \{[1, 15], [16, 20], [21, 29], [30, 42]\}$ is such a partition for the set of intervals shown in Figure 5. We shall map each element of $\mathcal{I}$ to an integer corresponding to its rank in the natural order. We denote by $R_{\mathcal{I}}(\mathcal{X})$ the reduction of $\mathcal{X}$ by the partition $\mathcal{I}$. The reduction has as many variables as $\mathcal{X}$ (equation 2 below) but the domains are replaced with the set of intervals in $\mathcal{I}$ that overlap with the corresponding variable in $\mathcal{X}$ (equation 3 below). Observe that the domains remain intervals after the reduction.

$$R_{\mathcal{I}}(\mathcal{X}) = \{X'_1, \ldots, X'_{|\mathcal{X}|}\}. \tag{2}$$
$$\forall X'_i \in R_{\mathcal{I}}(\mathcal{X}), \; \mathcal{D}(X'_i) = \{I \mid I \in \mathcal{I} \; \& \; \mathcal{D}(X_i) \cap I \neq \emptyset\}. \tag{3}$$

For instance, the set of intervals depicted in Figure 5 can be reduced to the set shown in Figure 4, where each element in $\mathcal{I}$ is mapped to an integer in $[1, 4]$.

**Theorem 4** *If $\mathcal{I}$ is a partition of $[1, \lambda]$ such that every element of $\mathcal{I}$ is a crest of occ, then* $\mathbf{ME}(\mathcal{X}) = \mathbf{ME}(R_{\mathcal{I}}(\mathcal{X}))$.

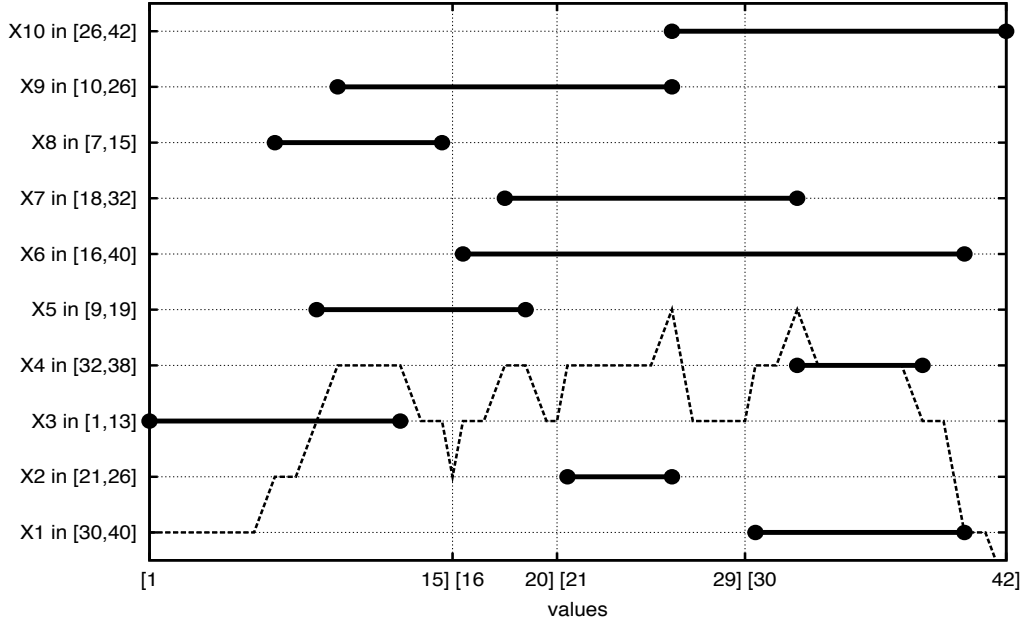

Figure 5: Some intervals and the corresponding *occ* function.

**Proof.** First, we show that for any optimal solution $s \in \mathbf{ME}(\mathcal{X})$, we can produce a solution $s' \in \mathbf{ME}(R_{\mathcal{I}}(\mathcal{X}))$ that has at least as many equalities as $s$. Indeed, for any value $a$, consider every variable $X$ assigned to this value, that is, such that $s[X] = a$. Let $I \in \mathcal{I}$ be the crest containing $a$, by definition we have $I \in \mathcal{D}(X')$. Therefore we can assign all these variables to the same value $I$.

Now we show the opposite, that is, given a solution to the reduced problem, one can build a solution to the original problem with at least as many equalities. The key observation is that, for a given crest $[a, b]$, all intervals overlapping with $[a, b]$ have a common value. Indeed, suppose that this is not the case, that is, there exists $[c_1, d_1]$ and $[c_2, d_2]$ both overlapping with $[a, b]$ and such that $d_1 < c_2$. Then $occ(d_1) > occ(d_1 + \frac{1}{2})$ and similarly $occ(c_2 - \frac{1}{2}) < occ(c_2)$. However, since $a \le d_1 < c_2 \le b$, $[a, b]$ would not satisfy the conditions for being a crest, hence a contradiction. Therefore, for a given crest $I$, and for every variable $X'$ such that $s'[X'] = I$, we can assign $X$ to this common value, hence obtaining as many equalities. □

We show that this transformation can be achieved in $\mathcal{O}(n \log n)$ steps. We once again use the derivative of the occurrence function ($\delta_{occ}$), however, defined on $Q$ rather than $[1, \lambda]$:

$$\delta_{occ}(v) \leftarrow (|\{i \mid \min(X_i) = v\}| - |\{i \mid \max(X_i) = v - \frac{1}{2}\}|).$$

Moreover, we can compute it in $\mathcal{O}(n \log n)$ steps as shown in Algorithm 4. We first compute the non-null values of $\delta_{occ}$ by looping through each variable $X \in \mathcal{X}$ (Line 1). We use the same data structure as for Algorithm 1, hence the complexity of this step is $\mathcal{O}(n \log n)$. Next, we create the partition into crests by going through the derivative once and identifying the inflection points. The variable *polarity* (Line 3) is used to keep track of the evolution of the function *occ*. The decreasing phases are denoted by *polarity = neg* whilst the increasing phases correspond to *polarity = pos*. We know that a value $v$ is the end of a crest interval when the variable *polarity* switches from *neg* to *pos*. Clearly, the number of elements in $\delta_{occ}$ is bounded by $2n$. Recall that the list data structure is sorted. Therefore, going through the values $\delta_{occ}(v)$ in increasing order of $v$ can be done in linear time, hence the overall $\mathcal{O}(n \log n)$ worst-case time complexity.

---

**Algorithm 4:** Computing a partition into crests.

---

**Data**: A set of variables: $\mathcal{X}$
**Result**: $\mathcal{I}$
$\delta_{occ} \leftarrow \emptyset$;
1 **foreach** $X \in \mathcal{X}$ **do**
    $\delta_{occ}(\min(X)) \leftarrow \delta_{occ}(\min(X)) + 1$;
    $\delta_{occ}(\max(X) + \frac{1}{2}) \leftarrow \delta_{occ}(\max(X) + \frac{1}{2}) - 1$;

$\mathcal{I} \leftarrow \emptyset$;
$min \leftarrow max \leftarrow 1$;
2 **while** $\delta_{occ} \neq \emptyset$ **do**
3     $polarity \leftarrow pos$;
    $k = 1$;
    **repeat**
        pick and remove the first element $(a, k)$ of $\delta_{occ}$;
        $max \leftarrow round(a) - 1$;
        **if** *polarity = pos* & $k < 0$ **then** $polarity \leftarrow neg$;
    **until** *polarity = pos or $k < 0$* ;
    add $[min, max]$ to $\mathcal{I}$;
    $min \leftarrow max + 1$;
return $\mathcal{I}$

---

Therefore, we can replace every crest by a single value at the preprocessing stage and then run Algorithm 3. Moreover, observe that the number of crests is bounded by $n$, since each needs at least one interval to start and one interval to end. Thus we obtain the following theorem, where $n$ stands for the number of variables, $\lambda$ for the number of distinct values, and $m$ for the sum of all domain sizes.

**Theorem 5** *Enforcing* RC *on* SoftAllEqual$_G^{min}$ *can be achieved in* $\mathcal{O}(\min(\lambda^2, n^2)nm)$ *steps.*

**Proof.** If $\lambda \leq n$ then one can achieve range consistency by iteratively calling Algorithm 3 after assigning each of the $\mathcal{O}(m)$ unit assignments ($(X, v) \ \forall X \in \mathcal{X}, v \in \mathcal{D}(X)$). The resulting complexity is $\mathcal{O}(n\lambda^2)m$ (see Theorem 3, the term $\lambda^3$ is absorbed by $n\lambda^2$ due to $\lambda \leq n$).

Otherwise, if $\lambda > n$, the same procedure is used, but after applying the reformulation described in Algorithm 4. The complexity of the Algorithm 4 is $\mathcal{O}(n \log n)$, and since after the reformulation we have $\lambda = \mathcal{O}(n)$, the resulting complexity is $\mathcal{O}(n^3 m)$. $\square$

# 7. Approximation Algorithm

We have completed the taxonomy of soft global constraints introduced in Section 3. However, in this section and in the rest of the paper we refine our analysis of the problem of maximising the number of pairs of variables sharing a value, that is, SOFTALLEQUAL$_G^{min}$-OPTIMISATION (Definition 10).

Given a solution $s$ over a set of variable $\mathcal{X}$, we denote by $obj(s)$ the number of equalities in $\mathcal{X}$.

$$obj(s) = |\{\{i, j\} \mid s[X_i] = s[X[j] \,\&\, i \neq j\}|.$$

Furthermore, we shall denote as $s^*$ and $obj(s^*)$ an optimal solution and the number of equalities in this solution, respectively. We first study a natural greedy algorithm for approximating the maximum number of equalities in a set of variables (Algorithm 5). This algorithm picks the value that occurs in the largest number of domains, and assigns as many variables as possible to this value (this can be achieved in $\mathcal{O}(m)$). Then it recursively repeats the process on the resulting sub-problem until all variables are assigned (at most $\mathcal{O}(n)$ times). We show that, surprisingly, this straightforward algorithm approximates the maximum number of equalities with a factor of $\frac{1}{2}$ in the worst case. Moreover, it can be implemented to run in $\mathcal{O}(m)$ amortised time. We use the following data structures[3]:

- $var : \Lambda \mapsto 2^{\mathcal{X}}$ maps every value $v$ to the set of variables whose domains contain $v$.

- $occ : \Lambda \mapsto \mathbb{N}$ maps every value $v$ to the number of variables whose domains contain $v$.

- $val : \mathbb{N} \mapsto 2^{\Lambda}$ maps every integer $i \in [0..n]$ to the set of values appearing in exactly $i$ domains.

These data structures are initialised in Lines 1, 2 and 3 of Algorithm 5, respectively. Then, Algorithm 6 recursively chooses the value with largest number of occurrences (Line 2), makes the corresponding assignments (Line 7) while updating the current state of the data structures (Loop 3).

---

**Algorithm 5:** Computing a lower bound on the maximum number of equalities.

**Data**: A set of variables: $\mathcal{X}$
**Result**: An integer $E$ such that $obj(s^*)/2 \leq E \leq obj(s^*)$

**1** $var(v) \leftarrow \emptyset, \; \forall v \in \bigcup_{X \in \mathcal{X}} \mathcal{D}(X)$;
    **foreach** $X \in \mathcal{X}$ **do**
        **foreach** $v \in \mathcal{D}(X)$ **do**
             add $X$ to $var(v)$;

**2** $occ(v) \leftarrow |var(v)|, \; \forall v \in \bigcup_{X \in \mathcal{X}} \mathcal{D}(X)$;
**3** $val(k) \leftarrow \emptyset, \; \forall k \in [0..n]$;
    **foreach** $v \in \bigcup_{X \in \mathcal{X}} \mathcal{D}(X)$ **do**
         add $v$ to $val(|var(v)|)$;

    return `AssignAndRecurse`$(var, val, occ, n)$;

---

**Theorem 6 (Algorithm Correctness)** *Algorithm 5 approximates the optimal satisfying assignment of the* SOFTALLEQUAL$_G$ *constraint within a factor of* $\frac{1}{2}$ *and - provided that the data-structure for representing domains respects some assumptions - runs in* $\mathcal{O}(m)$.

---

3. We describe these structures at a lower level in the subsequent proof of complexity.

---

**Algorithm 6:** procedure `AssignAndRecurse` of Algorithm 5.

---

**Data**: A mapping: $var : \Lambda \mapsto 2^{\mathcal{X}}$, A mapping: $val : \mathbb{N} \mapsto 2^{\Lambda}$, A mapping: $occ : \Lambda \mapsto [0..n]$, An integer: $k$

**1** **while** $val(k) = \emptyset$ **do** $k \leftarrow k - 1$;

    **if** $k \leq 1$ **then**

        | return 0;

    **else**

**2**          pick and remove any $v \in val(k)$;

**3**          **foreach** $X \in var(v)$ **do** **if** $v \in \mathcal{D}(X)$ **then**

             **foreach** $w \neq v \in \mathcal{D}(X)$ **do**

**4**                  remove $w$ from $val(occ(w))$;

**5**                  $occ(w) \leftarrow occ(w) - 1$;

**6**                  add $w$ to $val(occ(w))$;

**7**              assign $X$ with $v$;

         return $\frac{k(k-1)}{2}$+`AssignAndRecurse`$(var, val, k)$;

---

**Proof.** We first prove the correctness of the approximation ratio, the soundness of the algorithm and then the complexity of the algorithm.

*Approximation Factor.* We proceed using induction on the number of distinct values $\lambda$ in the current subproblem involving all unassigned variables. Let $s$ be the solution computed by Algorithm 5 and let $s^*$ be an optimal solution. We denote as $P(\lambda)$ the proposition "If there are no more than $\lambda$ values in the union of the domains of $\mathcal{X}$, then $obj(s) \geq obj(s^*)/2$". $P(1)$ implies that every unassigned variable can be assigned to a unique value $v$. Algorithm 6 therefore chooses this value and assigns all variables to it. In this case $obj(s) = obj(s^*)$.

Now we suppose that $P(\lambda)$ holds and we show that $P(\lambda + 1)$ also holds. Let the set of variables $\mathcal{X}$ of the problem be such that $|\bigcup_{X \in \mathcal{X}} \mathcal{D}(X)| = \lambda + 1$ and let $v$ be the first value chosen by Algorithm 6. We partition the variables into two subset $\mathcal{X}_v$ and $\bar{\mathcal{X}}_v$ depending on the presence of the value $v$ in their domains.

- $\mathcal{X}_v = \{X \in \mathcal{X} \mid v \in \mathcal{D}(X)\}$ is the set of variables whose domains contain $v$.

- $\bar{\mathcal{X}}_v = \mathcal{X} \setminus \mathcal{X}_v$ is the complementary set of variables which do not contain the value $v$.

Using these notations, we will partition the equalities into two subsets in order to count them. The first subset of equalities are those involving at least one variable in $\mathcal{X}_v$, the second subset are those restricted to variables in $\bar{\mathcal{X}}_v$.

We first compute a bound on the number of equalities that one can achieve on $\mathcal{X}$. Let $k = |\mathcal{X}_v|$, let $s_v^*$ be an optimal solution on $\bar{\mathcal{X}}_v$ and let $obj(s_v^*)$ be the number of equalities in $s_v^*$. For each variable $X \in \mathcal{X}_v$, given any value $w$ in $\mathcal{D}(X)$, there are no more than $k$ variables in $\mathcal{X}$ containing $w$. Indeed, $v$ was chosen for maximising this criterion and belongs to the domains of exactly $k$ variables. Therefore, there are at most $k(k-1)$ equalities that involve at least a variable in $\mathcal{X}_v$, since each one can be involved in at most $k - 1$ equalities, and there are $k$ of them. Consequently, on the set of variables $\mathcal{X}$, one can achieve at most $k(k-1) + obj(s_v^*)$ equalities.

On the other hand, Algorithm 6 assigns every variable in $\mathcal{X}_v$ to $v$ and therefore produces $k(k-1)/2$ equalities involving at least one variable in $\mathcal{X}_v$. Moreover, observe that since $v$ does not belong to any domain in $\bar{\mathcal{X}}_v$, the number of distinct values in $\bar{\mathcal{X}}_v$ is at most $\lambda$.

The induction hypothesis $P(\lambda)$ can therefore be used, hence we know that the number of equalities achieved by Algorithm 5 on the subset $\bar{\mathcal{X}}_v$ is at least $obj(s_v^*)/2$. Consequently, on the set of variables $\mathcal{X}$, Algorithm 5 achieves at least $k(k-1)/2 + obj(s_v^*)/2$ equalities.

Since the lower bound on the number of equalities achieved by the greedy algorithm is half of the upper bound computed above, we can conclude that if $P(\lambda + 1)$ holds, then $P(\lambda + 1)$ also holds.

*Correctness.* Here we show that the mappings $occ$ and $val$ are correctly updated in a call to Algorithm 6. The domain of a variable $X$ changes only when it is assigned to a value $v$ in Line 7. In that case, the occurrence of every value $w \in \mathcal{D}(X)$ such that $w \neq v$ is decreased by one when assigning $X$ to $v$. Indeed, for every such value $w$, $occ(w)$ is decremented and $w$ is removed from $val(occ(w) + 1)$ and added to $val(occ(w))$.

*Complexity.* Now we show that Algorithm 5 runs in $\mathcal{O}(m)$ steps under the following assumptions:

- The values are consecutive and taken from the set $\{1, \ldots, \lambda\}$.

- Assigning a variable to a value can be done in constant time.

- Checking membership of a value in a variable's domain can be done in constant time.

Notice that if the first assumption does not hold, one can rename values. However, it would require a further $\mathcal{O}(\lambda \log \lambda)$ time complexity to sort them, as well as $\mathcal{O}(n\lambda)$ to create a new set of domains.

For every $0 \leq k \leq n$, we use a doubly linked list to represent $val(k)$. Moreover we use a single array $index$ with $\lambda + 1$ elements to store the current position of every value $v$ in the list it appears in (observe that each value appears in exactly one list). To *add* a value $v$ in $val(k)$ we simply append it at the tail of the list and set its index to the previous length. To *remove* a value $v$ from $val(k)$, we delete the element at position $index[v]$ in $val(k)$. The total space complexity for this data-structure is therefore $\mathcal{O}(\lambda)$. For each value $v$, the set of variables $var(k)$ is implemented as a simple list, hence a $\mathcal{O}(m)$ space complexity. The mapping $occ(v)$ is represented as an array with one element per value, hence a $\mathcal{O}(\lambda)$ space complexity.

Initialising all three mappings is done in linear time since each addition requires only constant time. This step can therefore be achieved in $\mathcal{O}(m)$ steps. In Line 1 of Algorithm 6, $k$ can be decremented at most $n$ times in total, hence Line 2 is executed at most $n$ times in total.

Observe that no value is chosen more than once in Line 2. Moreover, the total space complexity of $var$ is $\mathcal{O}(m)$. Therefore, the total number of steps in Loop 3 is $\mathcal{O}(m)$.

Last, observe that no pair variable/value $(X, w)$ will be explored more than once in Lines 4, 5 and 6. Indeed, since $X$ is assigned to $v$ in Line 7, it will never pass the condition in Line 3 since subsequent chosen values will not be equal to $v$. The overall time complexity is thus in $\mathcal{O}(m)$.

□

**Theorem 7 (Tightness of the Approximation Ratio)** *The approximation factor of $\frac{1}{2}$ for Algorithm 5 is tight.*

**Proof.** Let $\{X_1, \ldots, X_4\}$ be a set of four variables with domains as follows:

$$X_1 \in \{a\}; \ X_2 \in \{b\}; \ X_3 \in \{a, c\}; \ X_4 \in \{b, c\}.$$

Every value appears in exactly two domains, hence Algorithm 5 can choose any value. We suppose that the value $c$ is chosen first. At this point no other value can contribute to an equality, hence Algorithm 5 returns 1. However, it is possible to achieve two equalities with the following solution: $X_1 = a, \ X_3 = a, \ X_2 = b, \ X_4 = b$. □

## 8. Tractable Class

In this section we explore further the connection between the SoftAllEqual$_G^{min}$ constraint and vertex matching. We showed earlier that the general case was linked to **3dMatching**. We now show that the particular case where no value appears in more than two domains solving the SoftAllEqual$_G$ constraint is equivalent to the vertex matching problem on general graphs, and therefore can be solved by a polynomial time algorithm. We shall then use this tractable class to show that SoftAllEqual$_G$ is NP-hard only if an unbounded number of values appear in more than two domains.

**Definition 11 (The VertexMatching Problem)**
Data: *An integer $K$, an undirected graph $G = (V, E)$.*
Question: *Does there exist $M \subseteq E$ such that $|M| \geq K$ and $\forall e_1, e_2 \in M$, $e_1$ and $e_2$ do not share a vertex.*

**Theorem 8 (Tractable Class of SoftAllEqual$_G^{min}$)** *If all triplets of variables $X, Y, Z \in \mathcal{X}$ are such that $\mathcal{D}(X) \cap \mathcal{D}(Y) \cap \mathcal{D}(Z) = \emptyset$ then finding an optimal satisfying assignment to SoftAllEqual$_G^{min}$ is in $P$.*

**Proof.** In order to solve this problem, we build a graph $G_{\mathcal{X}} = (V, E)$ with a vertex $x_i$ for each variable $X_i \in \mathcal{X}$, that is, $V = \{x_i \mid X_i \in \mathcal{X}\}$. Then for each pair $\{i, j\}$ such that $\mathcal{D}(X_i) \cap \mathcal{D}(X_j) \neq \emptyset$, we create an undirected edge $\{i, j\}$; let $E = \{\{i, j\} \mid i \neq j \ \& \ \mathcal{D}(X_i) \cap \mathcal{D}(X_j) \neq \emptyset\}$.

We first show that if there exists a matching of cardinality $K$, then there exists a solution with at least $K$ equalities. Let $M$ be a matching of cardinality $K$ of $G_{\mathcal{X}}$, for each edge $e = (i, j) \in M$ we assign $X_i$ and $X_j$ to any value $v \in \mathcal{D}(X_i) \cap \mathcal{D}(X_j)$ (by construction, we know that there exists such a value). Observe that no variable is considered twice since it would mean that two edges of the matching have a common vertex. The obtained solution therefore has at least $|M|$ equalities.

Now we show that if there exists a solution $S$ with $K$ equalities, then there exists a matching of cardinality $K$. Let $S$ be a solution, and let $M = \{\{i, j\} \mid S[X_i] = S[X_j]\}$. Observe that $M$ is a matching of $G_{\mathcal{X}}$. Indeed, suppose that two edges sharing a vertex (say $\{i, j\}, \{j, k\}$) are both in $M$. It follows that $S[X_i] = S[X_j] = S[X_k]$, however this is in contradiction with the hypothesis. We can therefore compute a solution $S$ maximising the number of equalities by computing a maximal matching in $G_{\mathcal{X}}$. □

This tractable class can be generalised by restricting the number of occurrences of values in the domains of variables. The notion of *heavy values* is key to this result.

**Definition 12 (Heavy Value)** *A heavy value is a value that occurs more than twice in the domains of the variables of the problem.*

**Theorem 9 (Tractable Class with Heavy Values)** *If the domain $\mathcal{D}(X_i)$ of each variable $X_i$ contains at most one heavy value then finding an optimal satisfying assignment of $\textsc{SoftAllEqual}_G^{min}$ is in $P$.*

**Proof.**  Consider a two stage algorithm. In the first stage, we explore every heavy value $w$ and assign $w$ to every variable whose domain contains it. Notice that no variable will be assigned twice. In the second stage, the CSP created by the domains of unassigned variables consists of only values having at most two occurrences, so we solve this CSP by transforming it to the matching problem as suggested in the proof of Theorem 8.

We show that there exists an optimal solution where each variable that can be assigned to a heavy value is assigned to this value. Let $s^*$ be an optimal solution and $w$ be a heavy value over a set $T$ of variables of cardinality $t$. We suppose that only $z < t$ of them are assigned to $w$ in $s^*$. Consider the solution $s'$ obtained by assigning all these $t$ variables to $w$: we add exactly $t(t-1)/2 - z(z-1)/2$ equalities. However, we potentially remove $t - z$ equalities since values other than $w$ do not appear more than twice. We therefore have $obj(s') - obj(s^*) \geq t^2 - 3t - z^2 + 3z$, which is non-negative for $t \geq 3$ and $z < t$. By iteratively applying this transformation, we obtain an optimal solution where each variable that can be assigned to a heavy value is assigned to this value. The first stage of the algorithm is thus correct. The second stage is correct by Theorem 8.  □

## 9. Parameterised Complexity

We further advance our analysis of the complexity of the $\textsc{SoftAllEqual}_G^{min}$ constraint by introducing a fixed-parameter tractable (FPT) algorithm with respect to the number of values. This result is important because it shows that the complexity of propagating this constraint grows only polynomially in the number of variables. It may therefore be possible to achieve AC at a reasonable computational cost even for a very large set of variables, provided that the total number of distinct values is relatively small.

We first show that the $\textsc{SoftAllEqual}_G^{min}$-OPTIMISATION problem is FPT with respect to the number of values $\lambda$. We use the tractable class introduced in the previous section to generalise this result, showing that the problem is FPT with respect to the number of *heavy* values occurring in domains containing two or more heavy values. We begin with a definition.

**Definition 13 (Solution from a Total Order)** *A solution $s_\prec$ is induced by a total order $\prec$ over the values if and only if*

$$s[X] = v \;\Rightarrow\; \forall w \prec v, \; w \notin \mathcal{D}(X).$$

We now prove the following key lemma.

**Lemma 1** *There exists a total order $\prec$ over the set of values, such that the solution $s_\prec$ induced by $\prec$ is optimal.*

**Proof.** Let $s^*$ be an optimal solution, $v$ be a value, and $occ(s^*, v)$ be the number of variables assigned to $v$ in $s^*$. Moreover, let $\prec_{occ}$ be a total order such that values are ranked by decreasing number of occurrences ($occ(s^*, v)$) and ties are broken arbitrarily. We show that $\prec_{occ}$ induces $s^*$.

Consider, without loss of generality, a pair of values $v, w$ such that $v \prec_{occ} w$. By definition we have $occ(s^*, v) \geq occ(s^*, w)$. We suppose that the hypothesis is falsified and show that this leads to a contradiction. Suppose that there exists a variable $X$ such that $\{v, w\} \subseteq \mathcal{D}(X)$ and $s^*[X] = w$ (that is, $\prec_{occ}$ does not induce $s^*$). The objective value of the solution $s'$ such that $s'[X] = v$ and $s'[Y] = s^*[Y]$ $\forall y \neq x$ is given by: $obj(s') = obj(s^*) + occ(s^*, v) - (occ(s^*, w) - 1)$. Therefore, $obj(s') > obj(s^*)$. However, $s^*$ is optimal, hence this is a contradiction. □

An interesting consequence of Lemma 1 is that searching over the space of total orders on values is enough to compute an optimal solution. Moreover, the fixed-parameter tractability of the SoftAllEqual$_G^{min}$ constraint follows easily from the same lemma.

**Theorem 10 (FPT − number of values)** *Finding an optimal satisfying assignment of the* SoftAllEqual$_G^{min}$ *constraint is fixed-parameter tractable with respect to $\lambda$, the number of values in the domains of the constrained variables.*

**Proof.** Explore all possible $\lambda!$ permutations of values. For each permutation create a solution induced by this permutation. Compute the cost of this solution. Return the solution having the highest cost. According to Lemma 1, this solution is optimal. Creating an induced solution can be done by selecting for each domain the first value in the order. Clearly, this can be done in $\mathcal{O}(m)$. Computing the cost of the given solution can be done by computing the number of occurrences $occ(w)$ and then summing up $occ(w) * (occ(w) - 1)/2$ for all values $w$. Clearly, this can be done in $\mathcal{O}(m)$ as well. Hence the theorem follows. □

We can also derive the following corollary from Lemma 1:

**Corollary 2** *The number of optimal solutions of the CSP with the* SoftAllEqual$_G$ *is at most $\lambda!$.*

**Proof.** According to Lemma 1, each optimal solution is induced by an order over the values of the given problem. Clearly each order induces exactly one solution. Thus the number of optimal solution does not exceed the number of total orders which is at most $\lambda!$. □

Corollary 2 shows that the number of optimal solutions of the considered problem *does not* depend on the number of variables and they all can be explored by considering all possible orders of values. We believe this fact is interesting from the practical point of view because in essence it means that even enumerating all optimal solutions is *scalable* with respect to the number of variables. Moreover, we can show that SoftAllEqual$_G^{min}$ is fixed-parameter tractable with respect to the number of conflicting values, defined as follows.

**Definition 14 (Conflicting Value)** *A value $w$ of a given CSP is a* conflicting value *if and only if it is a heavy value and there is a domain $\mathcal{D}(X)$ that contains $w$ and another heavy value.*

**Theorem 11 (FPT − number of conflicting values)** *Let $k$ be the number of conflicting values of a CSP comprising only one* SOFTALLEQUAL$_G$ *constraint. Then the CSP can be solved in time* $\mathcal{O}(k!\sqrt{n}\lambda)$*, hence* SOFTALLEQUAL$_G^{min}$ *is fixed-parameter tractable with respect to $k$.*

**Proof.**  Consider all the permutations of the conflicting values. For each permutation perform the following two steps. In the first step for each variable $X$ where there are two or more conflicting values, remove all the conflicting values except the one which is the first in the order among the conflicting values of $\mathcal{D}(X)$ according to the given permutation. In the second stage we obtain a problem where each domain contains exactly one heavy value. Solve this problem polynomially by the algorithm provided in the proof of Theorem 9.

Let $s$ be the solution obtained by this algorithm. We show that this solution is optimal. Let $p^*$ be a permutation of *all* the values of the considered CSP so that the solution $s^*$ induced by $p^*$ has the highest possible cost. By Lemma 1, $s^*$ is an optimal solution. Let $p_1$ be the permutation of the conflicting values which is induced by $p^*$ and let $s_1$ be the solution obtained by the algorithm above with respect to $p_1$. By definition of $s$, $obj(s) \geq obj(s_1)$. We show that $obj(s_1) \geq obj(s^*)$ from which the optimality of $s$ immediately follows.

Observe that there is no $X$ such that $s^*[X] = w$ and $w$ was removed from $\mathcal{D}(X)$ in the first stage of the above algorithm where the permutation $p_1$ is considered. Indeed, $w$ can only be removed from $\mathcal{D}(X)$ if it is preceded in $p_1$ by a value $v \in \mathcal{D}(X)$. It follows that $w$ is also preceded in $p^*$ by $v$ and consequently $s^*(X) \neq w$. Thus $s^*$ is a solution of the CSP obtained as a result of the first stage. However $s_1$ is an *optimal* solution of that CSP by Theorem 9 and, consequently, $obj(s_1) \geq obj(s^*)$ as required.

Regarding the runtime, observe that the execution of the algorithm consists of $k!$ running an algorithm for finding the largest bipartite matching of the given graph. This graph has $n$ vertices (corresponding to the variables). Moreover, each edge is associated with a value and no two edges are associated with the same value (because when the matching applies each value has at most two occurrences). It follows that the graph has at most $\lambda$ edges. According to Micali and Vazirani (1980), the largest matching can be found in $O(\sqrt{n}\lambda)$, hence the upper bound.  □

This result shows that the complexity of propagating the SOFTALLEQUAL$_G^{min}$ constraint comes primarily from the number of (conflicting) values, whereas other factors, such as the number of variables, have little impact. Notice that detecting conflicting values can be done in linear time ($\mathcal{O}(m)$), by first counting occurrences of every value, then flagging any value with at least two occurrences as "heavy" and finally flagging heavy values as "conflicting" in every domain containing at least two of them.

Observe, moreover, the "exponential" part of this algorithm is based on the exploration of all possible orders over the given set of conflicting values. In fact the ordering relation between two values matters only if these values belong to a domain of the same variable. In other words consider a graph $H$ on values of the given CSP instance. Two values $a$ and $b$ are connected by an edge if and only if they belong to the domain of the same variable. Instead of considering all possible orders over the given set of values we may consider all possible ways of transforming the given graph into an acyclic digraph. The upper bound on the number of possible transformations is $2^{E(H)}$ where $E(H)$ is the number of edges of $H$. For sparse graphs such a bound is much more optimistic that $k!$. For example, if the average degree of a vertex is 4 then the number of considered partial orders is $2^{2k} = 4^k$.

## 10. Finding a Set of Similar or Diverse Solutions

Problems of similarity and diversity have a wide range of applications. Finding several diverse solutions can be used to sample the solution space, for instance for product recommendation (Shimazu, 2001), case-based reasoning (Smyth & McClave, 2001; Aha & Watson, 2001) or constraint elicitation (Bessière, Coletta, Koriche, & O'Sullivan, 2005; Gama, Camacho, Brazdil, Jorge, & Torgo, 2005).

Conversely, similarity is important for problems with a periodic aspect. For instance, a schedule or timetable may need to be computed on a weekly basis, but the constraints might change slightly from week to week. In this type of problems the regularity of the solutions, that is, the similarity between each week's solution, is a very valuable property (Groër, Golden, & Wasil, 2009).

Finally, finding similar solutions to a set of variants of a problem can be useful to find solutions that are robust to uncertainty. Suppose, for example, that we are to solve a Travelling Salesman Problem (TSP), however, the costs associated with a set of $k - 1$ links between pairs of cities are uncertain or variable over time. We would like to find an optimal, or near-optimal, route such that when the cost of traversing a link changes, a limited amount of re-routing is sufficient to obtain another near-optimal solution. For that purpose, one can build a similar structure as that pictured in Figure 6 by duplicating the TSP once per uncertain link, the last being the original formulation. In each duplicate, the cost of the corresponding link is then set to some expected upper bound. If we minimise the distance between solutions, we obtain a solution with good properties of robustness: if the cost associated with the $i^{\text{th}}$ link increases, the solution of the $i^{th}$ duplicate is a valid alternative avoiding this link (if it degrades the solution quality too much) whilst requiring a small amount of re-routing.

We therefore want to find a set of $k$ solutions — either pairwise similar or different — to a set of $k$ problems, distinct or not. A heuristic method was introduced to solve the problem of finding $k$ solutions of a constraint network, such that the minimum (resp. maximum) distance between all pairs of solutions is maximum (resp. minimum) by Hebrard, Hnich, O'Sullivan, and Walsh (2005). Since reasoning on the maximum minimum distance is NP-hard (Frances & Litman, 1997), it was proposed to use the sum of the Hamming distances instead. In this section, we first formally define the notion of Hamming distance between variables and between solutions. Next, we show that the constraints studied in this paper can help achieve AC and RC in polynomial time for respectively maximising and minimising the sum of pairwise distances between solutions to a set of problem instances.

### 10.1 Hamming Distance:

The Hamming distance between the instantiation of two variables $X$ and $Y$ is defined as follows:

$$\Delta_h(X, Y) = \begin{cases} 1 & \text{iff } X \neq Y \\ 0 & \text{otherwise} \end{cases}$$

$$
\begin{aligned}
P_1 : &\quad (X_1^1,\ X_2^1,\ X_3^1,\ \ldots,\ X_n^1) \\
P_2 : &\quad (X_1^2,\ X_2^2,\ X_3^2,\ \ldots,\ X_n^2) \\
\ldots & \\
P_k : &\quad (X_1^k,\ X_2^k,\ X_3^k,\ \ldots,\ X_n^k)
\end{aligned}
$$

Figure 6: The problem $P$, duplicated $k$ times.

Whereas the Hamming distance between two solutions $s_i$ and $s_j$ (over the sets of variables $\{X_1^i, \ldots, X_n^i\}$ and $\{X_1^j, \ldots, X_n^j\}$, respectively) is defined as:

$$
\Delta_h(s_i, s_j) = \sum_{1 \le \ell \le n} \Delta_h(X_\ell^i, X_\ell^j)
$$

Given a problem $P$ with $n$ variables $\{X_1, \ldots, X_n\}$, we duplicate $P$ $k$ times, with identical constraints if we seek a set of diverse solutions to $P$, or altered constraints to model the expected scenarios if we seek for a set of similar solutions for some variations of $P$ (see Figure 6).

Then the objective to maximise or minimise is the sum of the pairwise distances between the (sub-)solutions of the duplicated problems:

$$
\sum_{1 \le i < j \le k} \Delta_h(s_i, s_j) \tag{4}
$$

## 10.2 Constraint Formulation:

The first approaches to this problem relied on heuristic methods (Hebrard et al., 2005; Hentenryck, Coffrin, & Gutkovich, 2009), It was also shown that when the problem $P$ allows it, knowledge compilation methods could efficiently solve this problem (Hadzic, Holland, & O'Sullivan, 2009).

Here we show that one can achieve arc or bound consistency for maximising this objective function. Whilst arc consistency is NP-hard for minimisation, bounds consistency can be achieved in polynomial time both for minimisation and maximisation. First, we decompose the objective function described previously (Equation 4) using the SOFTALLEQUAL$_G^{min}$ or SOFTALLEQUAL$_G^{max}$ constraints for optimising, respectively, solution similarity or diversity. Then we shall see that achieving AC (resp. BC) on this decomposition is equivalent to achieving AC (resp. BC) on the global constraint defined by bounding the objective.

Remember that each row in Figure 6 represents a duplicate of the original set of variables $\{X_1, \ldots, X_n\}$. The objective function is defined as the sum of the Hamming distances between every pair of rows. However, consider now Figure 6 by vertical slices. Each column corresponds to the set of duplicates $\{X_i^j \mid 1 \le j \le k\}$ of an original original variable $X_i$. One can compute the contribution of this set of variables to the sum of Hamming distances between pairs of rows as the number of pairwise disequalities in the set: $|\{\{j, k\} \mid X_i^j \ne X_i^k \ \& \ j < k\}|$. Notice that this is precisely the definition of the cost to minimise (resp. maximise) in SOFTALLEQUAL$_G^{min}$ (resp. SOFTALLEQUAL$_G^{max}$).

Therefore we can model the objective function as the constraint networks shown in Figure 7, respectively for minimisation and maximisation. Notice that, to simplify the model, we use the following, equivalent formulation for $\textsc{SoftAllEqual}_G^{max}$, rather than Definition 3:

$$\textsc{SoftAllEqual}_G^{max}(\{X_1, \ldots, X_n\}, N) \Leftrightarrow N \leq |\{\{i, j\} \mid X_i \neq X_j \ \& \ i < j\}|.$$

minimise $\qquad \sum_{1 \leq i \leq n} N_i \quad$ subject to

$\forall 1 \leq i \leq n \quad \textsc{SoftAllEqual}_G^{min}(X_i^1, \ldots, X_i^k, N_i)$

maximise $\qquad \sum_{1 \leq i \leq n} N_i \quad$ subject to

$\forall 1 \leq i \leq n \quad \textsc{SoftAllEqual}_G^{max}(X_i^1, \ldots, X_i^k, N_i)$

Figure 7: A constraint network that minimises (resp. maximises) the sum of distances between pairs of solutions to $k$ vectors of variables.

Second, notice that the constraint networks depicted in Figure 7 are such that no two constraints share more than one variable, and there is no Berge-cycle (Berge, 1970) in the constraint hypergraph, that is, a sequence $C_1, X_1, C_2, \ldots, X_k, C_{k+1}$ such that:

- $X_1, \ldots, X_k$ are distinct variables,

- $C_1, \ldots, C_{k+1}$ are distinct constraints,

- $k \geq 2$ and $C_1 = C_{k+1}$,

- $X_i$ is in the scope of $C_i$ and $C_{i+1}$.

Indeed, the $\textsc{SoftAllEqual}$ constraints do not share any variable, and the overlap with the sum constraint is limited to a single variable with each $\textsc{SoftAllEqual}$. The constraint hypergraph is therefore Berge-acyclic, and in such constraint networks it was shown that propagating AC is sufficient to filter all globally inconsistent values (Janssen & Vilarem, 1988; Jégou, 1991).

Therefore, when every constraint in this network is AC (resp. RC), the network is globally arc consistent (resp. globally range consistent). We can view these two constraint networks as two global constraints, respectively $\mathcal{CN}_{\text{div}}$ and $\mathcal{CN}_{\text{sim}}$, over the set variables $\{X_i^j \mid 1 \leq i \leq n, \ 1 \leq j \leq k\}$ and a variable $N$ to represent the objective:

$$\mathcal{CN}_{\text{div}}(\{X_i^j \mid 1 \leq i \leq n, \ 1 \leq j \leq k\}, N) \Leftrightarrow$$
$$N \leq \sum_{1 \leq i \leq n} N_i \ \& \ \forall 1 \leq i \leq n \ \textsc{SoftAllEqual}_G^{max}(X_i^1, \ldots, X_i^k, N_i)$$

$$\mathcal{CN}_{\text{sim}}(\{X_i^j \mid 1 \leq i \leq n, \ 1 \leq j \leq k\}, N) \Leftrightarrow$$
$$N \geq \sum_{1 \leq i \leq n} N_i \ \& \ \forall 1 \leq i \leq n \ \textsc{SoftAllEqual}_G^{min}(X_i^1, \ldots, X_i^k, N_i)$$

Theorems 12 and 13 (where $m = \sum_{1 \leq i \leq n} |\mathcal{D}(X_i^1)|$ denotes the sum of the domain sizes in *one* copy of the problem) follow from, respectively, (van Hoeve, 2004) and Theorem 5:

**Theorem 12** *Enforcing* AC *on* $\mathcal{CN}_{\text{div}}(\{X_i^j \mid 1 \leq i \leq n, \ 1 \leq j \leq k\}, N)$ *can be achieved in* $\mathcal{O}(k^2m)$ *steps.*

**Proof.** Since the constraint network equivalent to $\mathcal{CN}_{\text{div}}$ is Berge-acyclic, we know that it is AC iff every constraint in the decomposition is AC. Moreover, we describe a filtering algorithm that requires only a bounded number of calls to the propagator of each constraint in the decomposition.

We assume that no variable's domain is completely wiped out during the process. If it was the case, the process would be interrupted earlier (as soon as an inconsistency is detected while achieving AC on a component).

We introduce some terminology:

- *property (1)* denotes the fact that for all $1 \leq i \leq n$ the domains of the variables in $X_i^j$ are consistent with the upper bounds of $N_i$,

- *property (2)* denotes the fact that for all $1 \leq i \leq n$ the domains of the variables in $X_i^j$ are consistent with the lower bounds of $N_i$,

- *property (3)* denotes the fact that the sum constraint ($N \leq \sum_{1 \leq i \leq n} N_i$) is BC (or equivalently AC).

First, the domain of some variable $X_i^j$ for $1 \leq i \leq n$ and $1 \leq j \leq k$ might have changed, as well as the lower bound of $N$. A change on the upper bound of $N$ either results on an immediate failure, or bears no consequences.

1. For every $i \in [1..n]$, we update the upper bound of the variable $N_i$ by calling the procedure proposed by van Hoeve (2004) to find the maximum possible number of disequalities. Hence property (1) holds.

2. We achieve BC (equivalent to AC in this case) on the sum constraint. Notice that only the upper bound of $N$ and the lower bounds of $N_i$ for some $1 \leq i \leq n$ will be updated, therefore property (1) and (3) hold.

3. For every $i \in [1..n]$, we prune the domains of the variables in $\{X_i^j \mid 1 \leq j \leq k\}$ by calling the filtering procedure proposed by van Hoeve (2004). Since this domain reduction will not trigger any further changes in the bounds of $N_i$, we know property (1), (2) and (3) hold, hence $\mathcal{CN}_{\text{div}}$ is AC.

The first phase requires $O(\sum_{1 \leq i \leq n} k^2 |\mathcal{D}(X_i^1)|)$, that is $O(k^2m)$ steps. The second phase requires $O(n)$ steps. Finally, the third phase, like the first, requires $O(k^2m)$ steps. Hence an overall $O(k^2m)$ time complexity.

**Theorem 13** *Enforcing* RC *on* $\mathcal{CN}_{\text{sim}}(\{X_i^j \mid 1 \leq i \leq n, \ 1 \leq j \leq k\}, N)$ *can be achieved in* $\mathcal{O}(k^4m)$ *steps.*

**Proof.** This proof is very similar to that of Theorem 12, if we swap upper and lower bounds, and if we use the procedure described in Section 6 for phase (1) and (3).

The first phase requires $O(k^3 n)$ steps. The second phase requires $O(n)$ steps. Finally, the third phase requires $O(\sum_{1 \le i \le n} k^4 |\mathcal{D}(X_i^1)|)$, that is $O(k^4 m)$ steps. Hence an overall $O(k^4 m)$ time complexity.

## 11. Conclusion

In many applications we are concerned with stating constraints on the similarity and diversity amongst assignments to variables. To formulate such problems we can use soft variants of the well known ALLDIFFERENT and ALLEQUAL constraints. In this paper we considered the global constraints ALLDIFFERENT and ALLEQUAL, and their optimisation variants, SOFTALLDIFF and SOFTALLEQUAL, respectively. Furthermore, we considered two cost functions, based either on the Hamming distance to a satisfying assignment or on the number of violations on the decomposition graph. We have shown that the constraint ensuring an upper bound on the Hamming distance with a solution satisfying the ALLEQUAL constraint can be propagated efficiently, both for arc and bounds consistency. Then we have shown that, on the one hand, deciding the existence of an assignment minimising the number of violation in the decomposition graph of the ALLEQUAL constraint is NP-complete, hence propagating arc consistency on the constraint ensuring this property is NP-hard. On the other hand, propagating bounds consistency on the same constraint can be done in polynomial time. Moreover, we have shown that this problem is fixed parameter tractable in the number of distinct values of the problem. This work complements nicely some earlier results of Cohen et al. (2004) showing that the language of soft binary equality constraints was NP-complete, for as few as three distinct values in the domains. In this paper we have shown that the problem remains NP-complete even if the graph of soft binary equality constraints forms a clique, however, becomes polynomial if the number of values is bounded.

This paper therefore provides a comprehensive complexity analysis of achieving AC and BC on an important class of soft constraints of difference and equality. Interestingly, this taxonomy shows that enforcing equality is harder than enforcing difference.

## Acknowledgments

Hebrard, O'Sullivan and Razgon are supported by Science Foundation Ireland (Grant Number 05/IN/I886). Marx is supported in part by the ERC Advanced grant DMMCA, the Alexander von Humboldt Foundation, and the Hungarian National Research Fund (Grant Number OTKA 67651).